\documentclass[final,3p,times,twocolumn,authoryear]{elsarticle}
\usepackage{amssymb}
\usepackage{amsmath}
\usepackage{amssymb}
\usepackage{color}
\usepackage{booktabs} 
\usepackage{wrapfig}
\usepackage{tikz}
\usepackage{comment}
\usepackage{color}
\usepackage{multirow}
\usepackage{kotex}
\usepackage{subcaption}
\usepackage{adjustbox}
\usepackage{float}
\usepackage{arydshln}

\journal{Neural Networks}

\begin{document}

\begin{frontmatter}

%% Title, authors and addresses

%% use the tnoteref command within \title for footnotes;
%% use the tnotetext command for theassociated footnote;
%% use the fnref command within \author or \affiliation for footnotes;
%% use the fntext command for theassociated footnote;
%% use the corref command within \author for corresponding author footnotes;
%% use the cortext command for theassociated footnote;
%% use the ead command for the email address,
%% and the form \ead[url] for the home page:
%% \title{Title\tnoteref{label1}}
%% \tnotetext[label1]{}
%% \author{Name\corref{cor1}\fnref{label2}}
%% \ead{email address}
%% \ead[url]{home page}
%% \fntext[label2]{}
%% \cortext[cor1]{}
%% \affiliation{organization={},
%%            addressline={}, 
%%            city={},
%%            postcode={}, 
%%            state={},
%%            country={}}
%% \fntext[label3]{}

\title{FPANet: Frequency-based Video Demoir\'{e}ing using Frame-level Post Alignment}
%% use optional labels to link authors explicitly to addresses:
%% \author[label1,label2]{}
%% \affiliation[label1]{organization={},
%%             addressline={},
%%             city={},
%%             postcode={},
%%             state={},
%%             country={}}
%%
%% \affiliation[label2]{organization={},
%%             addressline={},
%%             city={},
%%             postcode={},
%%             state={},
%%             country={}}

\affiliation[inst1]{organization={Department of Artificial Intelligence},%Department and Organization
            addressline={Korea University}, 
            %city={Seoul},
            %postcode={02841}, 
            country={South Korea}}

\affiliation[inst2]{organization={Department of Computer Science and Engineering},%Department and Organization
            addressline={Korea University}, 
            %city={Seoul},
            %postcode={02841}, 
            country={South Korea}}

\affiliation[inst3]{organization={Graduate School of Culture Technology},%Department and Organization
            addressline={KAIST}, 
            %city={Daejeon},
            %postcode={34141}, 
            country={South Korea}}
            
\affiliation[inst4]{organization={LG Display Research Center},
            %city={Seoul},
            %postcode={07795}, 
            country={South Korea}}

\author[inst1]{Gyeongrok Oh}
\author[inst1]{Sungjune Kim}
\author[inst4]{Heon Gu}
\author[inst3]{Sang Ho Yoon}
\author[inst2]{Jinkyu Kim*}
\author[inst1]{Sangpil Kim*}

\begin{abstract}
Moiré patterns, created by the interference between overlapping grid patterns in the pixel space, degrade the visual quality of images and videos. Therefore, removing such patterns~(demoiréing) is crucial, yet remains a challenge due to their complexities in sizes and distortions. Conventional methods mainly tackle this task by only exploiting the spatial domain of the input images, limiting their capabilities in removing large-scale moiré patterns. Therefore, this work proposes FPANet, an image-video demoiréing network that learns filters in both frequency and spatial domains, improving the restoration quality by removing various sizes of moiré patterns. To further enhance, our model takes multiple consecutive frames, learning to extract frame-invariant content features and outputting better quality temporally consistent images.
We demonstrate the effectiveness of our proposed method with a publicly available large-scale dataset, observing that ours outperforms the state-of-the-art approaches in terms of image and video quality metrics and visual experience.

\end{abstract}

% %%Graphical abstract
% \begin{graphicalabstract}
% \includegraphics{grabs}
% \end{graphicalabstract}

%Research highlights
% \begin{highlights}
%     \item  We propose a novel method for robust sound-guided image manipulation by audio-visual weakly paired contrastive learning.
    
%     \item  We demonstrate that our learned embedding space is effective by extensive comparative experiments of the state-of-the-art text and sound-guided image manipulation.
    
%     \item  We show an in-depth analysis of our sound-guided image manipulation method.
% \end{highlights}


\end{frontmatter}
% \vspace{-3.2em}
\renewcommand*\footnoterule{}
\let\thefootnote\relax\footnote[0]{\scriptsize{$^*$ Corresponding authors:
Sangpil Kim (spk7@korea.ac.kr)} and Jinkyu Kim (jinkyukim@korea.ac.kr)}

%% \linenumbers

%% main text
\section{Introduction}\label{sec1}

\begin{figure}[t]%
    \centering
    \includegraphics[width=\linewidth]{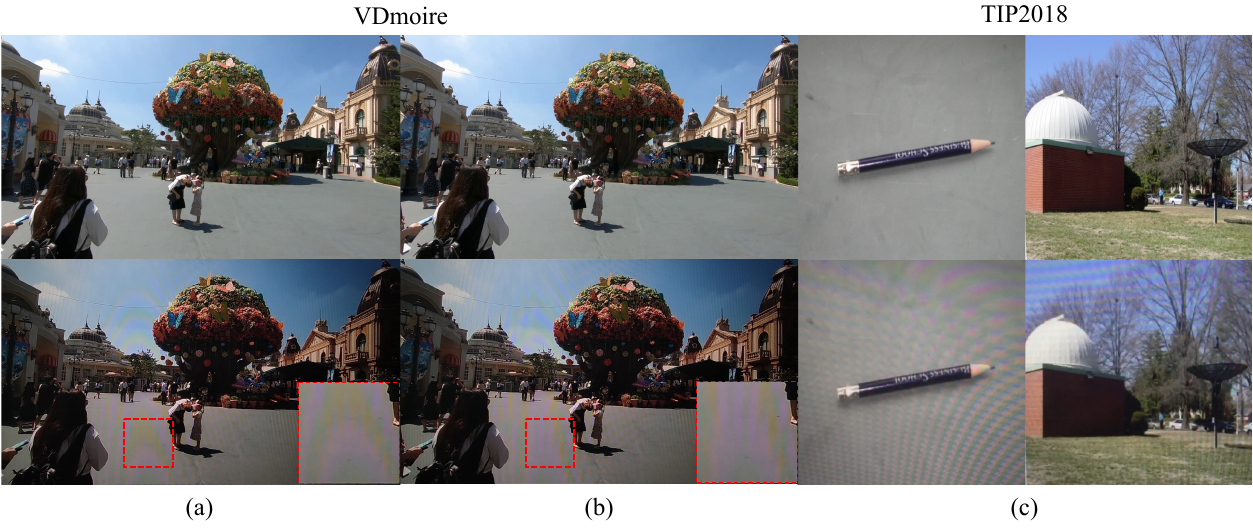}
    \caption{Examples of video frames or images that have moiré patterns as visual artifacts. Note that (a) and (b) are consecutive frames, which are extracted from the publicly available VDmoire dataset, while (c) is from the TIP2018 dataset. Target images are shown in the first row, and images with moiré patterns are shown at the bottom. For better visualization, we also provide magnified patches.}
    \label{fig:figure_input}
\end{figure}

Moiré patterns are commonly observed in visual contents, which are taken by ordinary digital cameras, capturing a screen of a digital display device. This is mainly due to frequency aliasing -- an interference between overlapping grid patterns, such as camera sensor grid and display pixel grid or textures on clothes. Such interference depends on the degree of overlap, resulting in diverse and complex patterns, including stripes, curves, and ripples, which are analytically infeasible (see Fig.~\ref{fig:figure_input}). They also significantly degrade the visual quality of images and videos, often causing severe color distortions of the original content. This makes it challenging to remove such moiré patterns and restore the original content. 

Learning-based approaches have been introduced to train a model to filter out visual artifacts and restore the original contents automatically. 
Recently, ConvNet-based hierarchical architectures have been explored to remove various sizes of moiré patterns in only spatial domain~\citep{sun2018moire, He_2019_ICCV, yu2022towards}. 
However, these approaches still suffer from the elimination of the large-scale moiré pattern and the recovery of color distortion.

\begin{figure}[t]
    \centering
    \includegraphics[width=\linewidth]{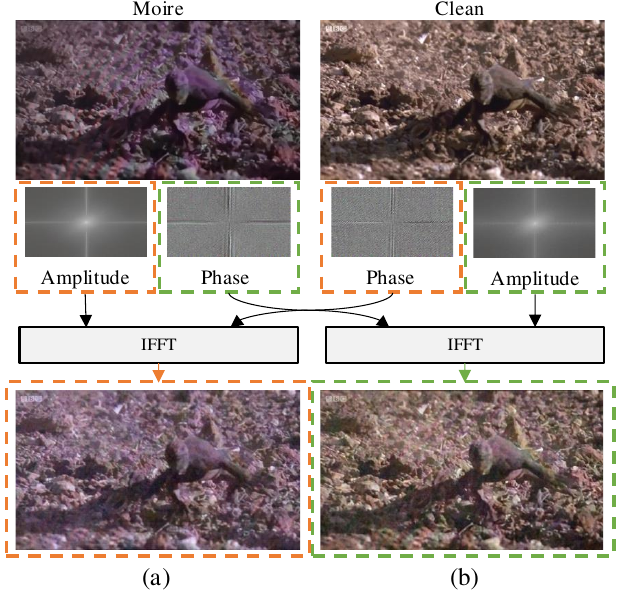}
    \caption{Visualization on the effect of amplitude and phase component over moiré patterns. The orange box generates a synthetic image combined with moiré image amplitude and clean image phase. The green box generates a synthetic image combined with the moiré image phase and clean image amplitude.}
    \label{fig:figure_freq}
\end{figure}

An alternative branch of research for demoiré focuses on leveraging frequency priors, such as the discrete wavelet domain~(DWT) or discrete Fourier transform~(DFT).
Recent works~\citep{liu2020wavelet, zheng2021learn, zheng2020image} that address artifacts across a wide range of frequencies caused by interference suggest that co-learning in the frequency domain is useful for handling moiré patterns of various sizes and shapes.
However, a frequency spectrum of the moiré pattern is often intermingled with the original contents. 
Moreover, the camera's Bayer filter mosaic is imbalanced over the RGB channel, producing different intensities of moiré patterns. 
Due to the intrinsic complexities of moiré patterns, a naive approach incorporating frequency domain into the conventional network is insufficient to completely separate the desired degraded frequency signal.

To address this issue, we advocate for leveraging amplitude and phase components separately. 
In our experiment (see Fig.~\ref{fig:figure_freq}), color distortion or degradation is often caused by the signal's amplitude component, while moiré patterns remain in the phase component. 
Thus, instead of directly utilizing frequency components as done in conventional approaches, encoding these components in a separate branch will accelerate the learning procedure.

\begin{figure*}[t]
    \centering
    \includegraphics[width=\textwidth]{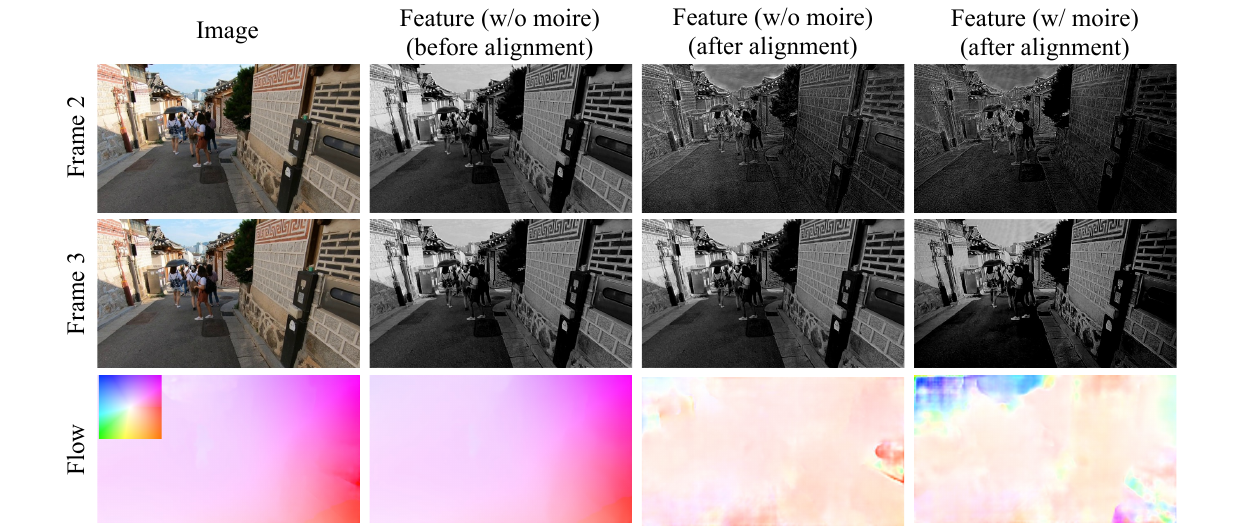}
    \caption{Example for misalignment caused by moiré patterns. Aligned features between reference frame~(Frame 2) and target frame~(Frame 3) with or without moiré patterns are listed in the first row. To measure the accuracy for alignment, we calculate optical flow using PWC-Net~\citep{sun2018pwc}, illustrating the relative motion by the color coding that indicates the motion vectors (direction and magnitude) with color intensity.}
    \label{fig:figure_align}
\end{figure*}

Thus, we propose a novel module called Frequency Selection Fusion (FSF), which first transforms the spatial information into the frequency domain spectrum using Fast Fourier Transform (FFT). Its amplitude and phase components are extracted and encoded separately. We further apply a selective fusion strategy to merge both components. Moreover, to maximize the representation power in the spatial domain, we adopt multi-scale architecture to improve restoring fine-grained details in the spatial domain.

This is only a part of the story. We further expand our work for video demoiréing tasks where we now take multiple consecutive frames, outputting temporally consistently restored images.
We argue that leveraging multiple frames with similar contents but slightly different moiré patterns is helpful to filter out such distortions, while augmented views of original contents are used to restore main content features.
Thus, a key module for video demoiréing models is aligning multiple consecutive frame inputs. Conventional aligning approaches~\citep{chan2021basicvsr, chan2022basicvsr++} used recurrent architectures (e.g. use bidirectional or uni-directional recurrent units followed by using an image warping technique), but they are prone to accumulate misalignment errors in a long sequence input, resulting in poor restoration performance. Enlarging receptive fields by applying deformable convolution~\citep{dai2017deformable, tian2020tdan, wang2019edvr, dai2022video} is an alternative way to align features of multiple consecutive frames. However, aligning features without removing moiré patterns often yields a large misalignment error, as shown in Fig.~\ref{fig:figure_align}. Thus, to address this issue, we propose an improved alignment module called Post Align Module (PAM). Unlike existing approaches that use an alignment module in the early stage (and are separated from the main network), we apply such an alignment module in the multiple intermediate stages where more distortions are getting removed. This allows the alignment module to be robust against moiré distortions.

Our proposed FPANet effectively integrates two significant components within a unified network for video demoiréing. First, we effectively remove moiré patterns with amplitude and phase components under the significant observation of the frequency domain. Second, we precisely encode temporal cues of the neighboring frames considering the underlying misalignment from the undesired artifacts.
To demonstrate the effectiveness of our methods, we conducted numerous ablation studies and show competitive results in various evaluation metrics, peak signal-to-ratio~(PSNR), structural similarity~(SSIM), Learned Perceptual Image Patch Similarity (LPIPS), Frechet Video Distance (FVD), and Feature-SIMilarity (FSIM). To quantitatively evaluate FPANet, we compare the difference between ground truth clean images and estimated images. Furthermore, we observed that our proposed methods are effective in removing moiré patterns and reconstructing a fine-detailed image compared with previous state-of-the-art methods. Our main contributions are summarized as follows:

\begin{itemize}
    \item We propose a novel building module called Frequency Selection Fusion~(FSF), which consists of two modules: (i) Frequency Selection Module (FSM) and (ii) Cross Scale Fusion Module (CSFM). Based on an in-depth examination of moiré patterns, our FSF transforms the input into the frequency domain spectrum and operates on their amplitude and phase components separately to remove large-scale moiré patterns without creating undesired color artifacts. In our extensive experiments, we demonstrate that these modules are effective in removing moiré patterns without creating noticeable visual artifacts. 
    \item To deal with multiple consecutive frames for video demoiréing tasks, we introduce an advanced temporal feature alignment module (PAM) deployed in the multiple intermediate stages to remove various types of moiré distortions.
    \item We compare our model with current state-of-the-art approaches on a publicly available video demoiréing dataset called VDmoire, and ours outperforms existing approaches in various image and video quality metrics, such as PSNR, SSIM, LPIPS, FVD, and FSIM.
\end{itemize}

\section{Related Works}\label{sec2}

\subsection{Moiré Removal}
The interference between two similar signals creates moiré pattern. In particular, when taking a picture of a display, the moiré pattern is caused by the misalignment between the grid of the display and the camera filter. Moiré patterns that are not present in the original image can seriously degrade the image quality because of the shape, which looks like a ripple, ribbon, or stripe, and the color change.
With the advances of deep learning and neural network, several works \citep{sun2018moire, He_2019_ICCV, kim2020c3net, peng2024image, du2024fc3dnet, cheng2024image} deeply explore the architectural design to address moiré patterns of various sizes through the hierarchical pipeline. Among these, DMHN~\citep{cheng2024image} addresses the image hiding task by utilizing the moiré patterns through deep neural networks, enhancing the effectiveness of concealing and recovering hidden images. In addition, FHD$e^2$Net~\citep{he2020fhde} and  ESDNet~\citep{yu2022towards} are designed for demoiréing high-resolution images, with dimensions such as 3840 $\times$ 2160. FDNet~\citep{liu2020self} restores a clean image given a degraded image, aided by a defocused moiré image. Furthermore, MRGAN~\citep{yue2021unsupervised} proposes an unsupervised method leveraging Generative Adversarial Network~\citep{goodfellow2014generative} without using explicit moiré and clean image pairs.
Apart from the extensive investigations on spatial domain, the frequency domain is also widely adopted to handle large-scale moiré patterns through wavelet transform~\citep{liu2020wavelet} and implicit Discrete Cosine Transform~(DCT)~\citep{zheng2019implicit, zheng2021learn, dai2023adaptive}. However, these methods still suffer from removing moiré patterns clearly because they do not sufficiently leverage the property of moiré patterns in the Fourier domain.

As an extension of image demoiréing research, video demoiré is firstly introduced by~\citep{dai2022video}, by restoring videos using relation-based consistency loss. DTCENet \citep{liu2024video} utilizes the invertible framework for better video restoration quality.
However, since the moiré patterns not only desaturate the colors across the entire image but also obscure the recognition of object shapes, this method can not fully recover the original color and fine-grained details. A concurrent work, DTNet~\citep{xu2024direction}, solely leverages the manually selected spectrum based on basic directions obtained from the Discrete Cosine Transform~(DCT). Nevertheless, they still overlook compositional interaction of amplitude and phase in forming moiré patterns.

In this paper, we concentrate on the video demoiré that has been explored less in the former literature.
Our proposed methods are the first works in demoiré task to remove moiré texture and recover the original color in the frequency domain using amplitude and phase. In addition, we design Post Align Module~(PAM) to leverage the neighboring frames as auxiliary information without disruption.

\subsection{Learning in the Frequency Domain}
Recently, learning in the frequency domain has been widely studied in various fields~\citep{9022018,mao2021deep, chi2020fast, chi2019fast, yang2020fda, yu2022frequency, cai2021frequency, qin2021fcanet, rao2021global, suvorov2022resolution, zou2021sdwnet, zhou2022spatial, jiang2021focal}. Some of these methods~\citep{mao2021deep, chi2020fast, chi2019fast} applied the spectral block correspondence to the vanilla convolution block where they utilize each grid value, which is associated with the frequency components. This is advantageous for leveraging a much larger receptive field than conventional convolution operations. Also, FcaNet~\citep{qin2021fcanet} proposed frequency channel attention using the discrete cosine transform~(DCT) to compress each feature over pre-defined DCT bases. Cai~\textit{et al.}~\citep{cai2021frequency} suggests the usefulness of amplitude and phase information for generating photo-realistic images in the generative models. Given this, Focal Frequency Loss~\citep{jiang2021focal} is utilized as a regularizer for the generative models. 
Also, FSDGN~\citep{yu2022frequency} and Fourmer~\citep{zhou2023fourmer} show similar motivation on amplitude and phase information of diverse degradation factors~(such as haze and rain).
However, those architectural designs fail to fully exploit both information with unidirectional flow (i.e. frequency to spatial), and neglect a thorough investigation on moiré patterns.
% Also, FSDGN~\citep{yu2022frequency} uses amplitude information to guide restoring the original image with haze.
% However, they separate spatial and frequency branches with the unidirectional flow (i.e. frequency to spatial) that can not fully use both information.
LaMa~\citep{suvorov2022resolution} shows remarkable performance in the inpainting task with Fast Fourier Convolution~\citep{chi2020fast}, which has the advantage of recovering repetitive structures such as fences. We also follow this line of work where we exploit amplitude and phase components separately via our proposed Frequency Selection Module~(FSM) for the video demoiréing task.

\subsection{Multi-frame Encoding and Alignment}
Techniques for encoding multiple consecutive frames can mainly be categorized into two ways: (i) window-based and (ii) recurrent-based methods. Window-based approaches~\citep{wang2019edvr, liu2021adnet, li2020mucan, dudhane2022burst, tian2020tdan, chadha2020iseebetter}  often use deformable convolution to have enlarged receptive field followed by encoding concatenated multiple frames. For example, TDAN~\citep{tian2020tdan} predicts offsets of the convolution kernel to align multi-frame features, EDVR~\citep{wang2019edvr} used Pyramid Cascade Deformable~(PCD), which is the hierarchical architecture to facilitate precise offset prediction, which is widely adopted in the field of video restoration task~\citep{wang2019edvr, liu2021adnet, dudhane2022burst, dai2022video, tian2020tdan}.

Recurrent-based methods often utilize recurrent units for encoding a sequence of frames: e.g. BasicVSR~\citep {chan2021basicvsr} and BasicVSR++~\citep{chan2022basicvsr++} leverage bidirectional flow to align multiple frames over time. However, dealing with long-term dependency issues remains challenging with the recurrent architectures, and their aligning performance is sub-optimal. Thus, in this paper, we pursue to align features between neighboring frames without the interference of moiré patterns and computational burden. Thus, we present the Post Align Module~(PAM) that aligns nearby features more accurately, and it can be easily plugged into a conventional convolution block.

\section{Method}\label{sec3}

In this paper, we propose a novel video demoiréing method called FPANet (Frequency-based video demoiré using frame-level Post Alignment). Following existing work, our model also adopts encoder-decoder architecture, which encodes an input image with moiré to a high-level feature representation, retaining geometric structural information for restoring high-quality images. Instead of taking a single image $I_t$ at a certain timestep $t$, our model takes the previous $I_{t-1}$ and the next frame $I_{t+1}$  as well. Thus, it takes three consecutive frames as input: $\mathbf{I}=\{I_{t-1}, I_t, I_{t+1}\}$. Given this, our model is trained to produce a restored image $P_t$.

\begin{figure*}[t]
    \centering
    \includegraphics[width=\textwidth]{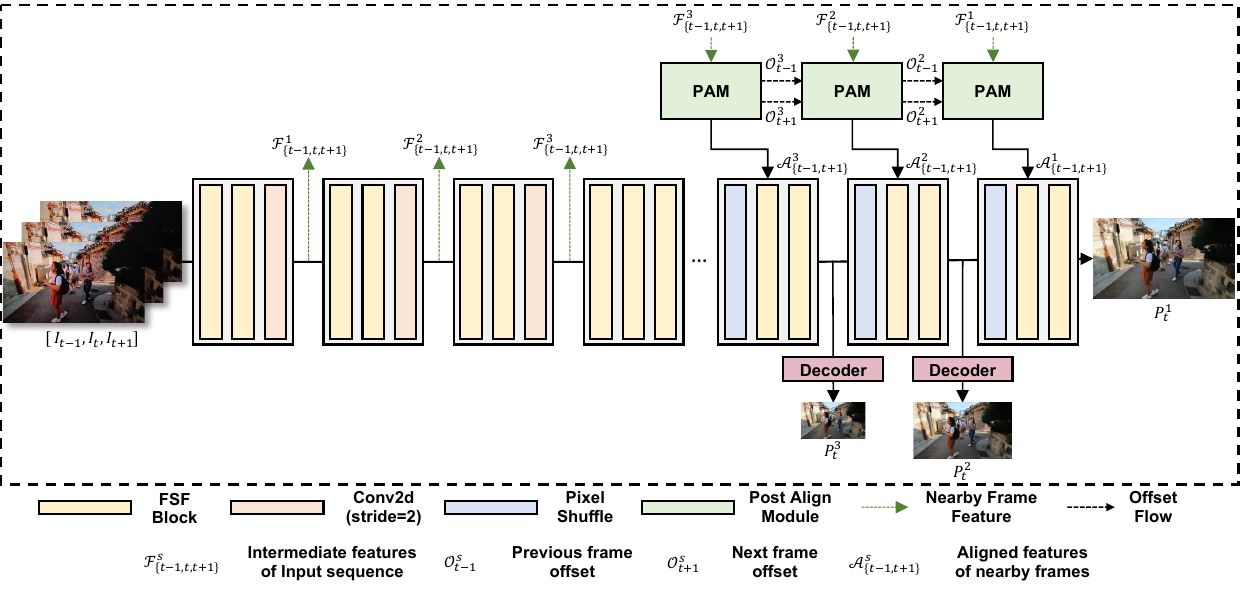}
    \caption{An overview of our proposed FPANet. FPANet is based on an encoder-decoder architecture design. Frequency Selection Fusion~(FSF) block is the core part of FPANet that is responsible for removing moiré patterns using frequency domain information. Also, Post Align Module~(PAM) is used for leveraging temporal information between nearby frames.}
    \label{fig:figure1}
\end{figure*}

As shown in Fig.~\ref{fig:figure1}, our model consists of the following two main components: (i) Frequency Selection Fusion (FSF) and (ii) Post Align Module (PAM). 
Our core module, FSF block, leverages spatial-frequency dual-domain features with two submodules: Frequency Selection Module (FSM) and Cross Scale Fusion Module (CSFM).
FSM first converts the incoming spatial-domain features into frequency-domain features, then its amplitude and phase components are separately processed to remove moiré patterns, retaining geometric structural information (Sect.~\ref{sec:fsm}). After processing frequency representation, multi-scale features for restoring both global and fine-grained features are incorporated in FSF using CSFM (Sect.~\ref{sec:csff}).

Lastly, PAM effectively aligns multiple consecutive frames with reducing visual artifacts (Sect~\ref{sec:pam}).

\begin{figure}[t]
    \centering
    \includegraphics[width=\linewidth]{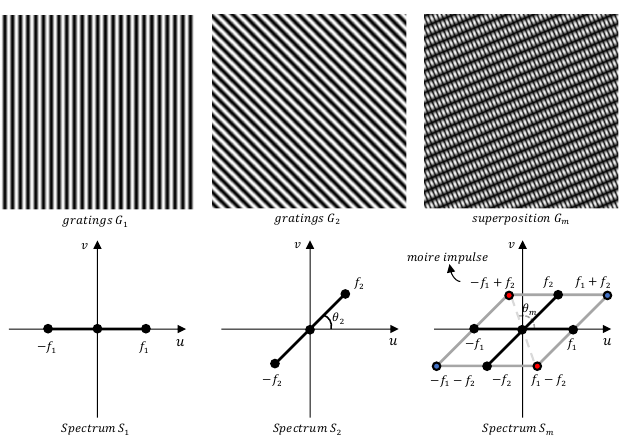}
    \caption{Example for moiré pattern caused by the superposition of two gratings and their spectrum. \textcolor{red}{Red} and \textcolor{blue}{blue} dot indicate the moiré impulse which is not appeared in the existing spectrum in the frequency domain}
    \label{fig:figure_moire}
\end{figure}

\begin{figure*}[t]
    \centering
    \includegraphics[width=\textwidth]{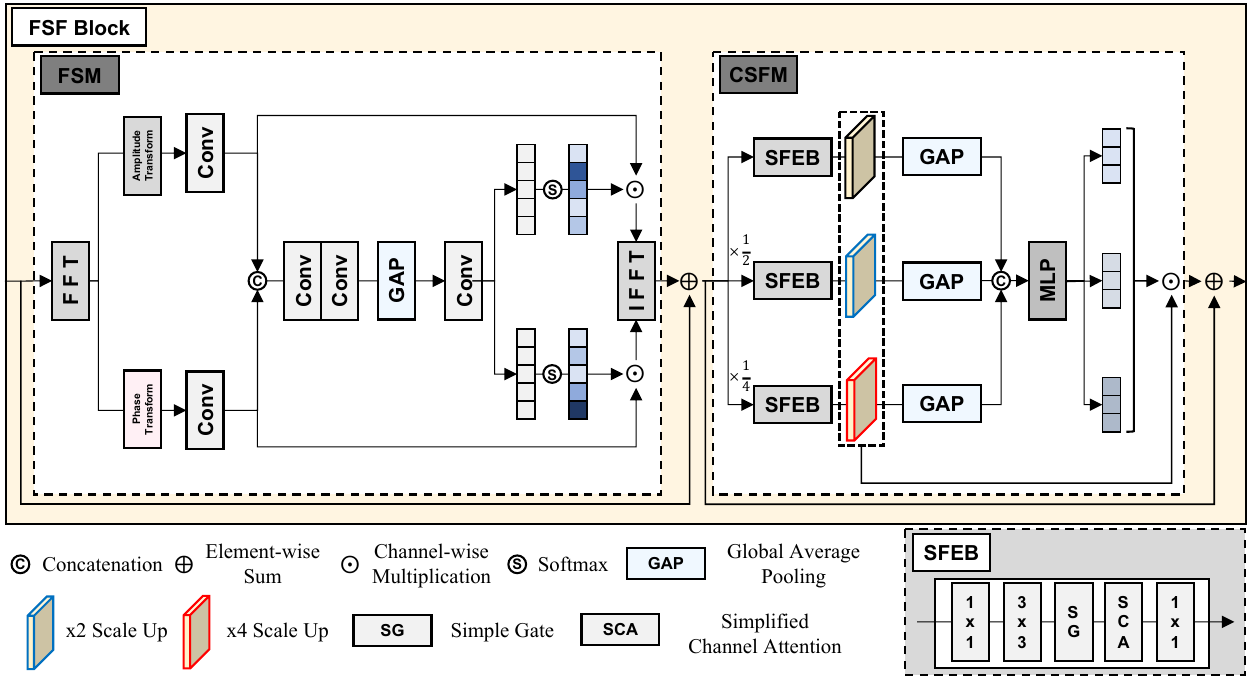}
    \caption{An overview of Frequency Selection Fusion Module, which consists of two main building blocks: (i) Frequency Selection Module (FSM) and (ii) Cross Scale Fusion Module (CSFM). FSM converts spatial information into frequency domain components and then encodes such frequency information to remove various types of moiré patterns effectively. CSFM processes a given spatial domain feature in a multi-scale processing manner.}
    \label{fig:figure2}
\end{figure*}

\subsection{Preliminary}\label{sec:preliminary}

\noindent\textbf{Moiré Pattern}
Moiré pattern, which arises when multiple repetitive structures overlap, is a well-known phenomenon that significantly degrades the quality of visual content.
To investigate the moiré pattern, we first consider two monochrome images which have periodic grating shapes. According to the~\citep{amidror2009theory, ning2022moirepose}, a simple cosinusoidal grating formulation follows as:
\begin{equation}
    \label{eq:gratings}
        G_i(x, y) = {1\over2}+{1\over2}cos(2\pi f_i(xcos\theta_{i}+ysin\theta_{i})),\quad i\in\{1,2\},
\end{equation}
where the $f$ indicates frequency and $\theta$ is the direction of periodic stripes~(illustrated in the $1^\text{st}$ row of Fig.~\ref{fig:figure_moire}). 
To obtain the superposition $G_m(x,y)$ as shown in Fig.~\ref{fig:figure_moire} (top right), we multiply two grating functions, $G_1\times G_2$:
\begin{equation}
\begin{split}
\label{eq:superposition}
    G_m(x,y) = ({1\over{2}}+{1\over 2} cos(2\pi f_1	[xcos\theta_1+ysin\theta_2]))\times \\
    ({1\over{2}}+{1\over{2}} cos(2\pi f_1[x cos\theta_2+y sin\theta_2])).
\end{split}
\end{equation}
Upon expanding Eq.~\ref{eq:superposition}, we can represent the newly generated impulse ($i.e.$, the moiré impulse) as follows: 
\begin{equation}
\label{eq:moire_impulse}
\begin{split}
     {1\over{8}} cos2\pi(f_1[xcos\theta_1+ysin\theta_1]-f_2[xcos\theta_2+ysin\theta_2]) + \\
     {1\over{8}} cos2\pi(f_1[xcos\theta_1+ysin\theta_1]+f_2[xcos\theta_2+ysin\theta_2]),
\end{split}
\end{equation}
where we present only two terms excluding the existing frequency impulse. Consequently, we can decompose moiré specific structures into frequency component $F_m(u,v)$, by applying a 2-dimensional Fourier Transform on Eq.~\ref{eq:moire_impulse}.

Additionally, in the frequency domain, we can easily explain the moiré pattern graphically based on Fourier approach.
To express the grating images as spectral representations in the frequency domain, we apply a 2-dimensional Fourier Transform and visualize the results $S_1(u,v)$ and $S_2(u,v)$ in $2^\text{nd}$ row of Fig.~\ref{fig:figure_moire}. 
According to the Fourier theorem~\citep{bracewell1986fourier}, the product of each grating function can be replaced by convolution of the spectral representation ($i.e.$, $S_m=$ $S_1$ $*$ $S_2$).
Thus, the convolution of the respective spectra generates two pairs of impulses and the original impulse as shown in spectrum $S_m$ of Fig.~\ref{fig:figure_moire} (bottom right). 
Note that we observe the additional frequency impulse ($f_1+f_2$ and $f_1-f_2$) that correspond to each term of Eq.~\ref{eq:moire_impulse} and visualize them as red and blue dots in Fig.~\ref{fig:figure_moire}. 
Thus, moiré patterns, which are not presented in original structures, not only exhibit visually distinct appearances but also demonstrate different properties in the frequency domain.

\noindent\textbf{Fourier Transform} 
One key component behind our model is encoding an input image in the frequency domain. Thus, we first explain how we convert a spatial domain feature $F(x,y)\in \mathbb{R}^{H\times W\times 3}$ into frequency components $\mathcal{F}(u,v)$. One common way to achieve this is via Discrete Fourier Transform (DFT), which decomposes spatial information (i.e. images) into frequency components using the following equation:
\begin{equation}
    \label{eq:eq2}
    \mathcal{F}(u,v)={\sum_{x=0}^{W-1}}{\sum_{y=0}^{H-1}}F(x, y)\cdot e^{-i2\pi({x\over W}u + {y \over H}v)}\, ,
\end{equation}
where $H$ and $W$ indicate the height and width of the image, respectively. This frequency information $\mathcal{F}(u,v)$ can further be decomposed into amplitude $\mathcal{F}^{\mathcal{A}}(u,v)$ and phase $\mathcal{F}^{\mathcal{P}}(u,v)$ components as follows: 
\begin{equation}
    \label{eq:amp-phase}
    \begin{split}
        \mathcal{F}^{\mathcal{A}}(u,v) &= \sqrt{\mathrm{Re}^2(u,v) + \mathrm{Im}^2(u, v)} \\    
        \mathcal{F}^{\mathcal{P}}(u,v) &= \mathrm{arctan}({\mathrm{Im}(u,v) / {\mathrm{Re}(u, v)}})
    \end{split}
\end{equation}
where $\mathrm{Re}(u,v)$ and $\mathrm{Im}(u, v)$ denote the real and imaginary value of complex Fourier coefficient, respectively (i.e. $\mathcal{F}(u,v)=\mathrm{Re}(u, v) + \mathrm{Im}(u, v)$).

\subsection{Frequency Selection Module (FSM)}\label{sec:fsm}
As shown in Fig.~\ref{fig:figure1}, our model consists of multiple Frequency Spatial Fusion (FSF) blocks in series, each of which contains two components: (i) Frequency Selection Module (FSM) and (ii) Cross Scale Fusion Module (CSFM). In this section, we explain the details of the FSM, and we will explain the CSFM in the next section. Note that $H_i$, $W_i$, and $C_i$ represent the height, width, and channel dimensions at $i$-th stage, respectively.
As shown in Fig.~\ref{fig:figure2} (left), the $i$-th FSM block takes as an input a (spatial domain) feature $F_{i}(x,y)\in \mathbb{R}^{H_{i} \times W_{i} \times C_{i}}$, outputting an encoded feature of the same dimension with a skip connection, i.e. $\texttt{FSM}(F_{i}(x,y))+F_{i}(x,y)$. 
The first step in the FSM block transforms the spatial features into frequency features $\mathcal{F}_i(u,v)\in \mathbb{R}^{H_{i}\times(\lfloor{W_{i}\over2}\rfloor+1)\times C_{i}}$ using 2D Fast Fourier Transform~(FFT)~\citep{5217220}. We only take half of the matrix to improve computational efficiency because 2D FFT produces a conjugate symmetric Hermitian matrix.

The resulting frequency features are then decomposed into amplitude $\mathcal{F}^{\mathcal{A}}_i(u,v)\in\mathbb{R}^{H_{i}\times(\lfloor{W_{i}\over2}\rfloor+1)\times C_{i}}$ and phase $\mathcal{F}^{\mathcal{P}}_i(u,v)\in\mathbb{R}^{H_{i}\times(\lfloor{W_{i}\over2}\rfloor+1)\times C_{i}}$ components using Eq.~\ref{eq:amp-phase}. Given these features, we apply a selective fusion strategy to filter out moiré patterns effectively, which generally depends on both amplitude and phase components. Inspired by prior works~\citep{li2019selective, zamir2022learning, fan2023lacn, liu2019adaptive}, we (channel-wise) concatenate both features, followed by $1\times1$ and $3\times3$ convolution layers in series to encode combined information. Then, we generate confidence maps ($\alpha_i\in\mathbb{R}^{C_i}$ and $\beta_i\in\mathbb{R}^{C_i}$) using a global average pooling (GAP) layer, another $1\times1$ convolution layer, and a softmax layer. These (channel-wise) confidence maps are then multiplied by $\mathcal{F}^{\mathcal{A}}_i(u,v)$ and $\mathcal{F}^{\mathcal{P}}_i(u,v)$ accordingly. Lastly, we apply 2-D Inverse Fast Fourier Transform~(IFFT) to transform frequency features into spatial features as follows:
\begin{equation}
\begin{split}
    \texttt{FSM}(F_{i}(x,y)) =
    \texttt{IFFT}(\texttt{Conv}(\alpha_i\odot\mathcal{F}^{\mathcal{A}}_i(u,v)), \\ \texttt{Conv}(\beta_i\odot\mathcal{F}^{\mathcal{P}}_i(u,v))))
\end{split}
\end{equation}
where $\odot$ is channel-wise multiplication.

\subsection{Cross Scale Fusion Module (CSFM)}\label{sec:csff}
As shown in Fig.~\ref{fig:figure2} (right), we further use Cross Scale Fusion Module (CSFM) to encode multi-scale features. As suggested by recent works~\citep{sun2018moire,zheng2021learn, zheng2020image, yu2022towards, zhang2022progressive}, such a multi-scale encoding is advantageous to restore fine-grained visual details. Following Yu~\textit{et al.}~\citep{yu2022towards}, we use the pyramid feature architecture to encode multi-scale features (i.e. original, $\times\, 1/2$ down-scaled, and $\times\,  1/4$ down-scaled features). 
However, we use a Simple Feature Extraction Block (SFEB) instead of dilated residual dense block due to improving the computational efficiency. 
In Fig.~\ref{fig:figure2} (bottom right), we illustrate SFEB, which consists of five layers including $1\times 1$ and $3 \times 3$ convolution layers, Simple Gate (SG), and Simplified Channel Attention (SCA)~\citep{chen2022simple}. After extracting multi-scale features from SFEB, we apply the Dynamic Fusion method~\citep{yu2022towards} in the same way to calibrate each scale feature into a unified feature map maintaining important information.

\begin{figure}[t]
    \centering
    \includegraphics[width=\linewidth]{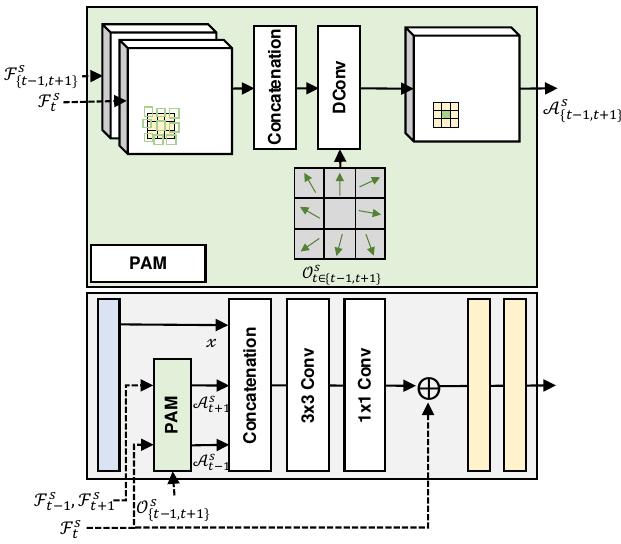}
    \caption{ Schematic for Post Align Module~(PAM). It is plugged into the decoder of the whole architecture to fuse multiple features. The dotted line indicates the skip connection from the encoder, and the solid line represents the straightforward flow in the module.}
    \label{fig:figure3}
\end{figure}

\begin{figure*}[t]
    \centering
    \includegraphics[width=\textwidth]{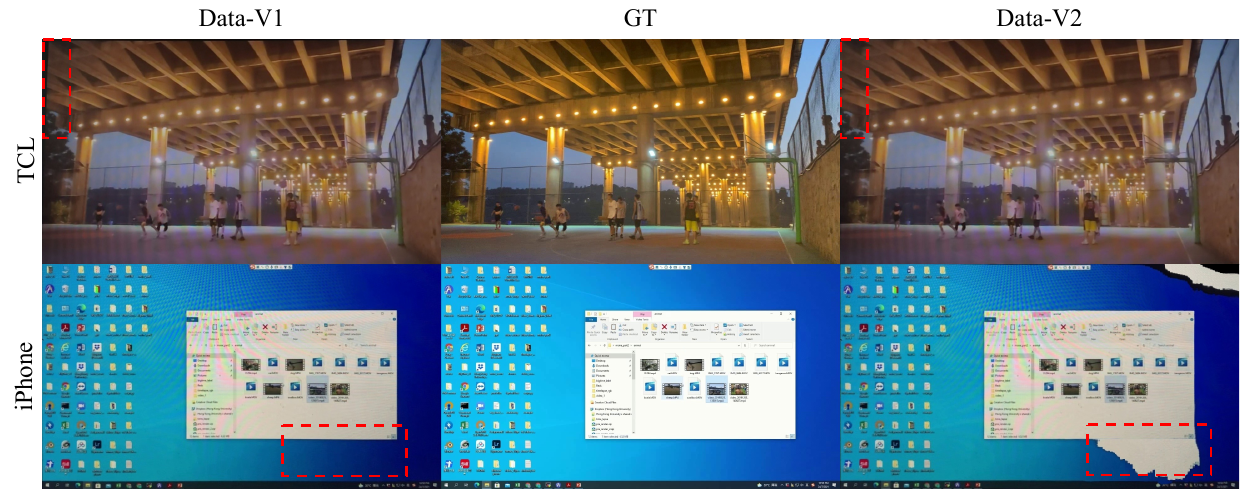}
    \caption 
    {Example of sampled frames in VDmoire~\citep{dai2022video} that is divided into two types of datasets conforming to different characteristics.
    The frame captured by two different types of device pairs~(TCL and iPhone) is listed in each row.
    In addition, each column represents an example of two versions of the dataset~(V1 and V2) before and after refinement, respectively, except for the second column, which indicates the ground truth.} 
    
    \label{fig:figure-dataset}
\end{figure*}

\subsection{Post Align Module (PAM)}\label{sec:pam}
Recent feature alignment methods~\citep{wang2019edvr} using Pyramid Cascade Deformable (PCD) or its deformation need to utilize a separated module and align each feature before reconstructing high-quality frames. However, moiré patterns are different in both size and properties compared with the other artifacts such as blur, rain streak, and Gaussian noise. Specifically, moiré patterns lead to severe color degradation and confused texture, which is an undifferentiated original texture.

Therefore, we propose the Post Align Module~(PAM) for aligning features between nearby frames without the interference of moiré patterns. As shown in Fig.~\ref{fig:figure3}, PAM is placed in the decoder per each stage. PAM uses features that are the output of each stage encoders over the sequence of frames $\mathbf{I}=\{I_{t-1}, I_t, I_{t+1}\}$ as an input. We define each input as $\mathcal{F}_{\{t-1, t, t+1\}}^{s}$, where $s$ denotes each stage~(e.g. $s \in \{1,2,3\}$). At the first stage of the PAM, we calculate learnable offset $\mathcal{O}_{\{t-1, t+1\}}$ using the output of the last stage of encoder $\mathcal{F}_{\{t-1, t, t+1\}}$:

\begin{equation}
        \mathcal{O}_{\{t-1, t+1\}} = \mathrm{Conv}([\mathcal{F}_{\{t-1, t+1\}}, \mathcal{F}_t]),
\end{equation}
where $\mathrm{Conv}(\cdot)$ represents the general convolution operation and $[\cdot,\cdot]$ indicates the concatenation operation.
Because the direction of the offset is identical and only changes magnitude regardless of the scales, we utilize the previously predicted learnable offset using bilinear interpolation. We formulate the following process: 

\begin{equation}
\begin{split}
        \mathcal{O}_{\{t-1, t+1\}}^{s} = 
        \mathrm{Conv}([\mathrm{Conv}([\mathcal{F}_{\{t-1, t+1\}}^{s}, \mathcal{F}_t^{s}]), \mathrm{Up}(\mathcal{O}_{\{t-1, t+1\}}^{s-1})]),
\end{split}
\end{equation}
where $\mathrm{Up}(\cdot)$ indicates the bilinear upscaling operation.
 
Formally, given learnable offset $\mathcal{O}_{\{t-1, t+1\}}$, we apply deformable convolution in order to align features according to the motion of objects. The overall PAM process is represented as:

\begin{equation}
    \mathcal{A}_{\{t-1, t+1\}}^{s} = \mathrm{Dconv}(\mathcal{F}_t^s, \mathcal{O}_{\{t-1, t+1\}}^{s}),
\end{equation}

where $\mathcal{A}$ represents the aligned feature and $\mathrm{Dconv}(\cdot)$ denotes the deformable convolution~\citep{dai2017deformable}. Last, obtained aligned features $\mathcal{A}_{\{t-1, t+1\}}$ and upsampled features $x$ from the previous block are processed using a conventional fusion method that consists of concatenation and convolution operation for leveraging each information. Note that the PAM can be easily plugged into the decoder to obtain the aligned features over neighboring frames.

\subsection{Loss Function}\label{sec:loss}
We train our model end-to-end using the following three loss functions: (i) Multi-scale $L_1$-based spatial domain loss, (ii) Multi-scale perceptual loss, and (iii) $L_1$-based frequency domain loss. Our Multi-scale $L_1$-based pixel-wise loss $\mathcal{L}_s$ quantifies the pixel-wise differences between the target image and the restored image, which is defined as follows:
\begin{equation}
    \mathcal{L}_s = \sum_t\|P_t-\hat{I}_t\|_{1}
\end{equation}
where $\hat{I}_t$ is the ground-truth (target) image without moiré patterns, and $P_t$ is the restored image by our proposed method. 

Following~\citep{yu2022towards}, we also use multi-scale perceptual loss by utilizing ImageNet pre-trained VGG19~\citep{simonyan2014very} network. We extract high-level feature representations from the intermediate layer and train a model to minimize the feature-level difference between the target and the predicted images. We define multi-scale perceptual loss as follows:
\begin{equation}
    \mathcal{L}_\text{vgg} = \sum_{t}\|\texttt{VGG}_{19}(P_t) - \texttt{VGG}_{19}(\hat{I}_t)\|_1
\end{equation}
where $\texttt{VGG}_{19}(\cdot)$ denotes the pre-trained network. 

Lastly, we also use a $L_1$-based frequency domain loss, which quantifies differences between frequency components (i.e. amplitude and phase) of the target and restored images. Our loss function is defined as follows:
\begin{equation}
    \mathcal{L}_\text{freq} = \sum_{t}\|\mathcal{F}^\mathcal{A}(P_t)-\mathcal{F}^\mathcal{A}(\hat{I}_t)\|_{1} + \|\mathcal{F}^\mathcal{P}(P_t)-\mathcal{F}^\mathcal{P}(\hat{I}_t)\|_{1}
\end{equation}
where $\mathcal{F}^\mathcal{A}$ and $\mathcal{F}^\mathcal{P}$ represent amplitude and phase components. Concretely, we use the following loss function $\mathcal{L}$:
\begin{equation}
    \mathcal{L} = \mathcal{L}_s + \lambda_\text{vgg}\mathcal{L}_\text{vgg} + \lambda_\text{freq}\mathcal{L}_\text{freq}
\end{equation}
where $\lambda_\text{vgg}$ and $\lambda_\text{freq}$ are the hyper-parameters to control the strength of each loss term.

\begin{table*}[hbt!]
    \begin{center}
    \caption{Quantitative comparison with the state-of-the-art image~\citep{ronneberger2015u, sun2018moire, yu2022towards, liu2020wavelet, zheng2020image} and video~\citep{dai2022video, liu2024video, xu2024direction} demoiréing approaches. We use the publicly available VDmoire dataset (which contains TCL-V1 and iPhone-V2 splits) for this experiment. Note that we use {\bf{bold}} to highlight the best scores in multi-frame experiments and \underline{underline} the best scores in single-frame experiments among different models.$\uparrow$ represents a higher score is better, while $\downarrow$ indicates a lower score is better. {\it{Abbr.}} Freq: use the frequency components, Multi: use the multiple consecutive frame inputs}
    \label{table:quantative-sota}
    \resizebox{\textwidth}{!}{
        \begin{tabular}{lcccccccc}\toprule
        \multirow{2}{*}{Method} &  \multirow{2}{*}{\parbox{.9cm}{\centering Freq}} & \multirow{2}{*}{\parbox{.9cm}{\centering Multi}} & \multicolumn{3}{c}{TCL-V1} & \multicolumn{3}{c}{iPhone-V2} \\ \cmidrule{4-9}
        & & &  PSNR$\uparrow$ & SSIM$\uparrow$ & LPIPS$\downarrow$ & PSNR$\uparrow$ & SSIM$\uparrow$ & LPIPS$\downarrow$\\\midrule
        DMCNN~\citep{sun2018moire} & -  & - & 20.321 & 0.703 & 0.321 & 21.816 & 0.749 & 0.496 \\
        UNet~\citep{ronneberger2015u} & -  & - & 20.348 & 0.720 & 0.225 & 21.678 & 0.790 & 0.338 \\
        WDNet~\citep{liu2020wavelet} & \checkmark & - & 20.576 & 0.697 & 0.234 & 23.971 & 0.834 & 0.205 \\
        MBCNN~\citep{zheng2020image} & \checkmark  & - & 21.534 & 0.740 & 0.260 & 24.060 & 0.849 & 0.211\\
        ESDNet~\citep{yu2022towards} & - & - & {\underline{\bf{22.026}}} & 0.734 & 0.199  & 25.064 & 0.853 & 0.165\\
        VDmoire~\citep{dai2022video}  & -  & \checkmark & 21.725 & 0.733 & 0.202 & 25.230 & 0.860 & 0.157\\
        DTCENet~\citep{liu2024video}  & -  & \checkmark & 21.881 & 0.744 & 0.181 & 25.381 & 0.876 & 0.148 \\
        DTNet~\citep{xu2024direction}  & \checkmark  & \checkmark & 21.846 & 0.734 & 0.178 & 25.317 & 0.846 & 0.164 \\
        \midrule
        Ours ({\it single frame only}) & \checkmark & - & 21.577 & \underline{0.772} & \underline{0.189} & \underline{25.215} & \underline{0.875} & \underline{0.157} \\
        Ours  & \checkmark & \checkmark & 21.953 & {\bf{0.784}} & {\bf{0.173}} & {\bf{25.446}} & {\bf{0.883}} & {\bf{0.146}} \\\bottomrule
    \end{tabular}}
\end{center}
\end{table*}

\section{Experiments}\label{sec4}
\subsection{Experimental Setup}
\noindent\textbf{Implementation Details} We train our model end-to-end with AdamW \citep{loshchilov2017decoupled} optimizer with the initial learning rate set to $10^{-3}$ and $\beta_1 = 0.9,\beta_2=0.999$ for 80 epochs.
We use cyclic cosine annealing learning rate schedule~\citep{loshchilov2016sgdr} that enables partial warm restart optimization, generally improving the convergence rate in gradient-based optimization. 
To train the model on the video demoiré task, we conduct single-frame and multi-frame experiments with the following settings. 
Specifically, at the multi-frame experiments, we randomly sample three consecutive frames with batch size $8$ and crop a 384$\times$384 patch for training. 
In the single-frame experiments, we randomly sample a single frame and leverage it repetitively instead of consecutive frames. To get the final result in test time, we utilize the original resolution for all datasets followed by~\citep{dai2022video}.
Note that we perform a grid search to find a better hyperparameter set using a mini version dataset~($\approx$20\%): we set $\lambda_\text{vgg}$ and $\lambda_\text{freq}$ to $0.1$.
Moreover, we set the number of FSF blocks at the encoding and decoding stage to [2, 2, 4] for each scale. The entire architecture also contains 12 FSF blocks between the encoder and decoder.

\noindent\textbf{Evaluation Metrics} We use the following three evaluation metrics: (1) Peak Signal to Noise Ratio (PSNR), (2) Structural Similarity index (SSIM) \citep{1284395}, and (3) Learned Perceptual Image Patch Similarity (LPIPS) \citep{zhang2018unreasonable}. These metrics are widely used in the image demoiréing task (as well as other image restoration tasks) to quantify the quality of generated outputs. PSNR quantifies pixel-level similarity, while SSIM~\citep{1284395} uses structural information (from pixel intensities, luminance, and contrasts), providing a more human-like perception metric. Lastly, LPIPS~\citep{zhang2018unreasonable} measures perceptual similarity by comparing high-level visual representations from pre-trained networks (e.g., ImageNet \citep{5206848} pre-trained).

\subsection{Dataset}
To evaluate the effectiveness of our proposed methods, we use the following publicly available dataset, i.e. VDmoire~\citep{dai2022video} dataset. This provides a video demoiréing dataset, which provides 290 source videos and corresponding videos with moiré patterns. To obtain such pairs, the 720p (1080$\times$720) source videos are displayed on the MacBook Pro display (or Huipu v270 display). At the same time, a hand-held camera (iPhoneXR or TCL20 pro camera) captures the screen to create moiré patterns in the recorded frames. Further, to reduce the effect of the misaligned frame correspondences, they estimate the homography matrix using the RANSAC algorithm to align two frames. 
Despite these efforts, errors of spatial alignment between the captured frame and ground truth still remain (see first row in Fig.~\ref{fig:figure-dataset}).
To handle this problem, they present a refined new dataset using optical flow. But, this dataset also has the problem of showing distortion caused by inaccurate optical flow~(see second row in Fig.~\ref{fig:figure-dataset}). To compare ours with the former state-of-the-art methods in diverse settings, we utilize two different datasets~(TCL-V1 and iPhone-V2).
Note that we call TCL-V1 and iPhone-V2 based on the cameras used (i.e. TCL20 pro camera and iPhone XR camera).

\begin{table}
\begin{center}
% \vspace{-2.5em}
\caption{Quantitative comparison in terms of FVD~\citep{unterthiner2018towards} and FSIM~\citep{zhang2011fsim} metrics to further analyze the quality of generated videos (or image frames). Note that lower FVD and higher FSIM scores are better.}
\label{table:video_comp}
 \resizebox{\linewidth}{!}{
        \begin{tabular}{@{}lcc@{}}
            \toprule
            Method  & FVD$\downarrow$ & FSIM$\uparrow$ \\\midrule
            DMCNN~\citep{sun2018moire}   & 992.86 & 0.857 \\
            UNet~\citep{ronneberger2015u}   & 928.81 & 0.878 \\
            WDNet~\citep{liu2020wavelet}  & 697.28 & 0.895 \\
            MBCNN~\citep{zheng2020image}   & 694.28 & 0.857 \\
            VDmoire~\citep{dai2022video}   & 697.09 & 0.911 \\
            ESDNet~\citep{yu2022towards} & 633.93 & 0.906 \\
            \midrule
            Ours   & \textbf{633.87} & \textbf{0.967}  \\
            \bottomrule
        \end{tabular}}
\end{center}
\end{table}

\begin{figure*}[!]
    \centering
    % \vspace{-1cm}
    \includegraphics[width=\textwidth]{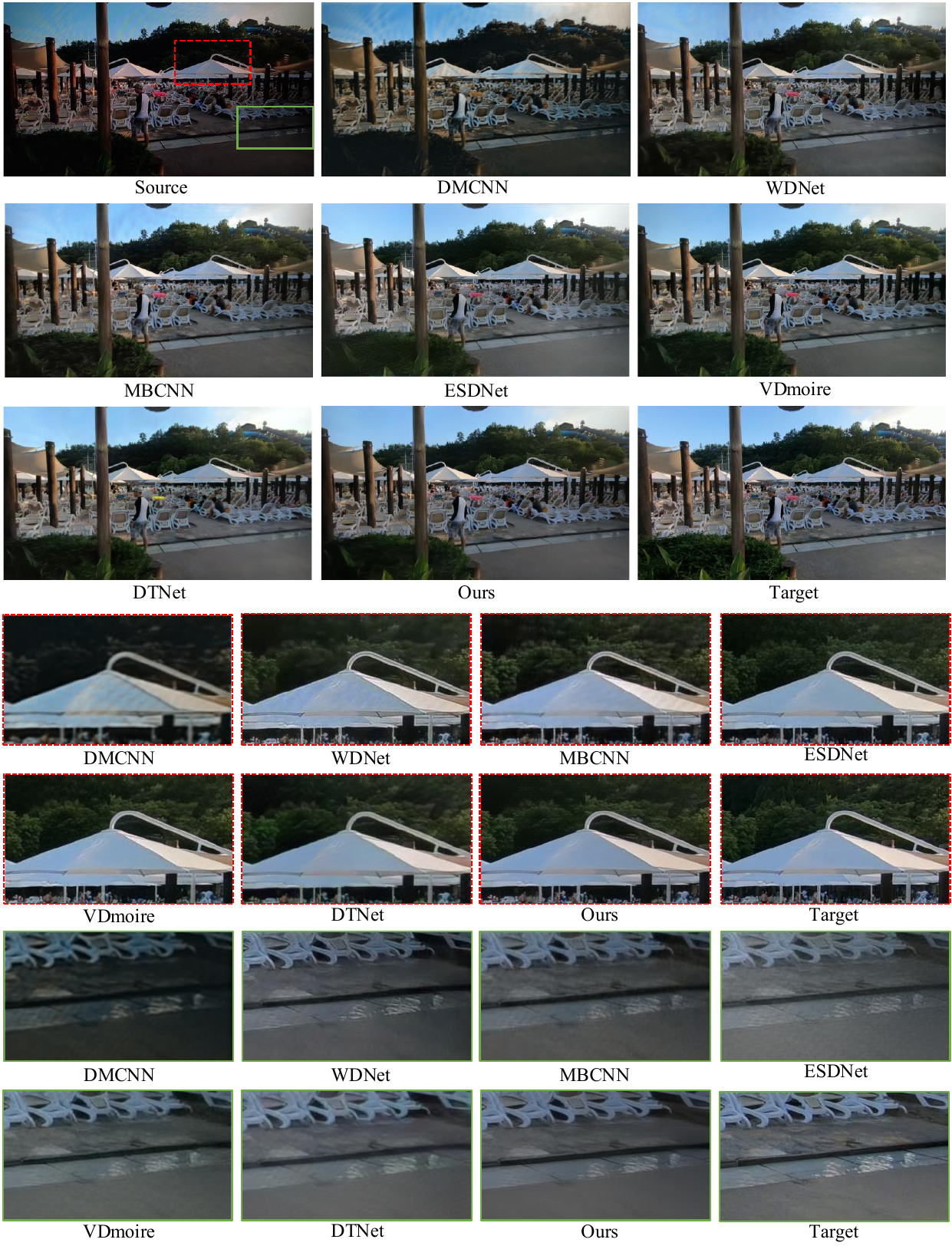}
    \caption{Qualitative comparison on the TCL-V1. The red and green boxes zoom in on frames to obviously compare the results.}
    \label{fig:figure-tcl-result}
\end{figure*}
% \clearpage

\begin{figure*}[!]
    \centering
    % \vspace{-1cm}
    \includegraphics[width=\textwidth]{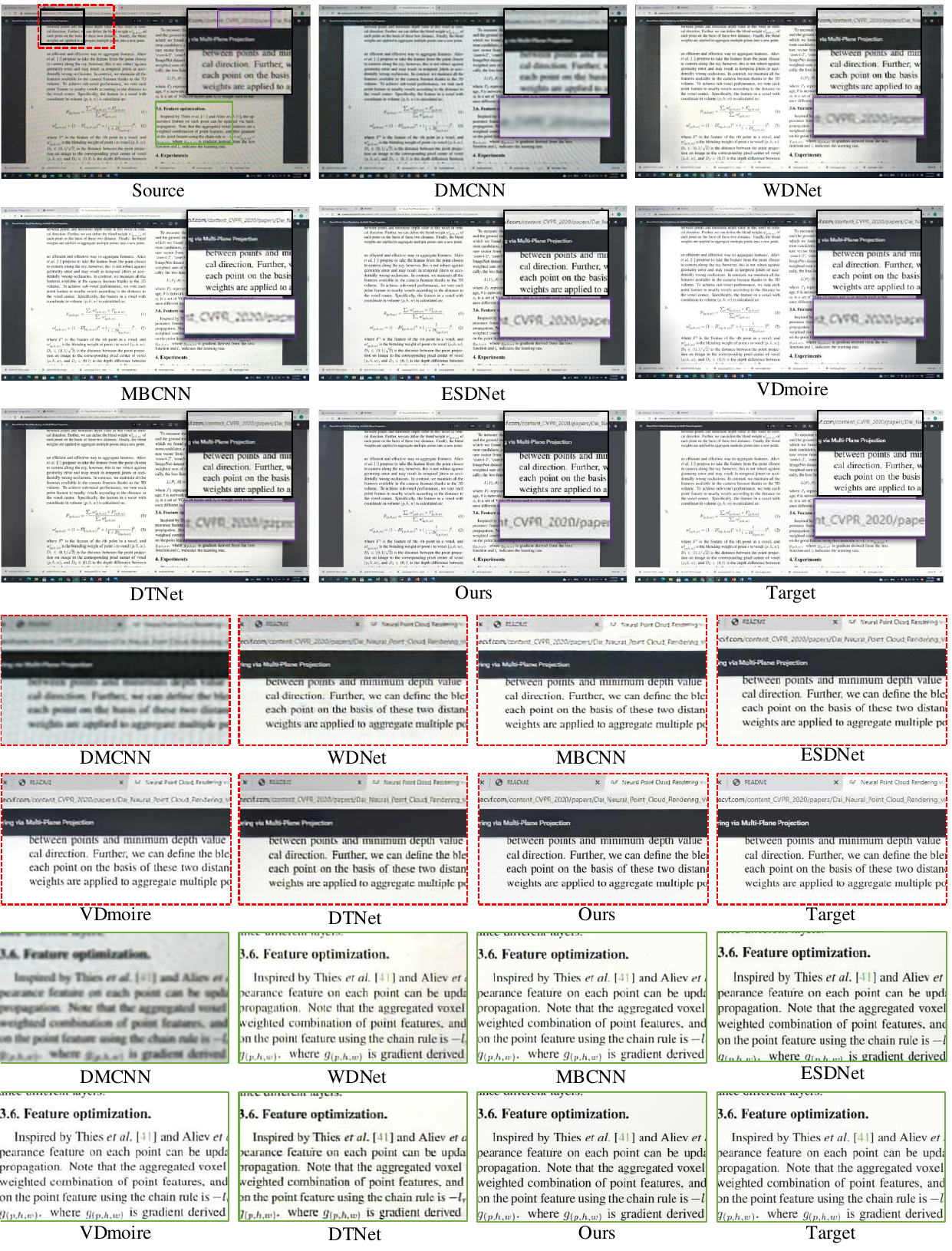}
    \caption{Qualitative comparison on the iPhone-V2. The red and green boxes zoom in on frames to obviously compare the results.}
    \label{fig:figure-iphone-result}
    % \vspace{-1.3em}
\end{figure*}
% \clearpage

\begin{figure*}[t]
    \centering
    \includegraphics[width=\textwidth]{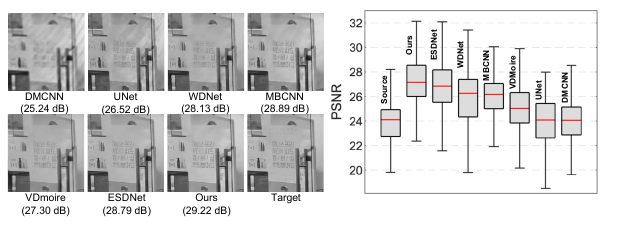}
    \caption{(left) Comparison of the luminance (Y channel in YCbCr color space) with state-of-the-art approaches (right) and their box plots of PSNR values.}
    \label{fig:figure-ychannel}
\end{figure*}

\begin{figure*}[t]
    \centering
    \includegraphics[width=\textwidth]{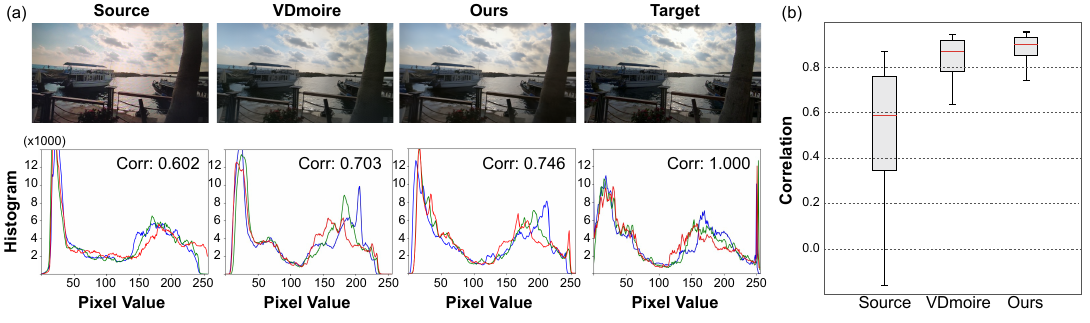}
    \caption{(a) We provide a sample of images (top) and their color histograms (bottom). Red, blue, and green lines indicate histograms for each RGB component: the x-axis represents the pixel values in [0, 255], while the y-axis denotes the number of pixels. (b) Box plots for correlation between the color distributions of the input and the target images.}
    \label{fig:figure-color-histogram}
\end{figure*}

\subsection{Quantitative Evaluation}
As we summarized in Table~\ref{table:quantative-sota}, we start by quantitatively comparing the quality of generated outputs with state-of-the-art approaches, including UNet~\citep{ronneberger2015u}, DMCNN~\citep{sun2018moire}, ESDNet~\citep{yu2022towards}, WDNet~\citep{liu2020wavelet}, MBCNN~\citep{zheng2020image}, and VDmoire~\citep{dai2022video}. Note that VDmoire~\citep{dai2022video} and ours utilize multi-frame inputs, while others are based only on single-frame image input. 
Also, we use the VDmoire dataset~\citep{dai2022video} for this evaluation, and the same hyper-parameters commonly used for image demoiréing, such as patch size, are used in this experiment for a fair comparison, while we use the default values for the other hyper-parameters.

We observe in Table 1 that our proposed method generally outperforms the other approaches in all metrics (except PSNR on TCL-V1 data) over other state-of-the-art approaches, both image- and video-based. Specifically, compared with the state-of-the-art video demoiréing methods, ours shows significant performance gain in PSNR, SSIM, and LPIPS metrics. Such a gain is also apparent in other perceptual evaluation metrics, i.e. SSIM and LPIPS. This indicates that ours produces more realistic and high perceptual quality. Moreover, amplitude and phase components enhance the efficacy of moiré pattern removal in comparison to the DTNet, which solely employs the Discrete Cosine Transform (DCT).

\begin{table*}[!t]
    \begin{center}
    \caption{We provide our ablation study to analyze the effect of individual building blocks: (i) Amplitude and Phase, (ii) FSM, (iii) CSFM, and (iv) PAM. We measure PSNR and SSIM for each combination of our four components with the number of parameters~(Params). Note that $\checkmark$ represents that the module is deployed.}
    \label{table:ablation-module}
    \resizebox{\textwidth}{!}{
         \begin{tabular}{lcccccc}
            \toprule
            Components & Model 1 & Model 2 & Model 3 & Model 4 & Model 5  & Model 6 (Ours)\\\midrule
            Amplitude \& Phase (Sect.~\ref{sec:fsm}) & -  &   -   & \checkmark  &\checkmark   & \checkmark & \checkmark    \\ 
            FSM (Sect.~\ref{sec:fsm}) & - &  - &   -   &\checkmark   &  \checkmark  &  \checkmark   \\  
            CSFM (Sect.~\ref{sec:csff}) & - & \checkmark  & \checkmark  &   -   &   \checkmark & \checkmark  \\
            PAM (Sect.~\ref{sec:pam}) & -  & \checkmark  & \checkmark  &\checkmark   &  - &\checkmark \\
            \midrule
            Params~(M)  & 37.98  & 59.95 & 58.44 & 58.51 & 64.97 &  68.89 \\ 
            \midrule
            PSNR$\uparrow$/SSIM$\uparrow$  & 20.51/0.75  & 21.00/0.76 & 21.42/0.77 & 21.73/0.77 & 21.13/0.77 & \textbf{21.95}/\textbf{0.78}  \\  
            \bottomrule
        \end{tabular}}
    \end{center}
\end{table*}

\begin{table}
\begin{center}
% \vspace{-2.5em}
\caption{We provide our ablation study to demonstrate the effect of each loss term ($\mathcal{L}_s$, $\mathcal{L}_\text{vgg}$, and $\mathcal{L}_\text{freq}$).}
\label{table:ablation_loss}
 \resizebox{\linewidth}{!}{%
    	      \begin{tabular}{@{}lccc@{}} \toprule
                {Loss function}  & PSNR$\uparrow$ & SSIM$\uparrow$ & LPIPS$\downarrow$ \\\midrule
                $\mathcal{L}_s$ & 21.653 & 0.770 &  0.240  \\
                $\mathcal{L}_s + \mathcal{L}_\text{vgg}$ &21.896  & 0.780 & 0.185  \\
                $\mathcal{L}_s + \mathcal{L}_\text{freq}$ & 21.883  &  0.781  & 0.207  \\\midrule
                $\mathcal{L}_s + \mathcal{L}_\text{vgg} + \mathcal{L}_\text{freq}$ (Ours) &  \textbf{21.953} &  \textbf{0.784} & \textbf{0.173} \\
            \bottomrule
        \end{tabular}}
\end{center}
\end{table}

\subsubsection{Temporal Consistency}
Consistent with the recent work~\citep{dai2022video}, we observe that using multiple consecutive input frames is helpful to improve the overall quality of video demoiréing (compare the bottom two rows vs. others in Table~\ref{table:quantative-sota}). Note that even without utilizing multi-frame image inputs, our proposed method still outperforms the other existing approaches, which justifies the effectiveness of using amplitude and phase components as in our Frequency Selection Module~(FSM) for removing moiré patterns. 

Following existing works~\citep{cho2022memory, skorokhodov2022stylegan}, we further use two metrics to analyze the quality of video outputs: FVD~\citep{unterthiner2018towards} and FSIM~\citep{zhang2011fsim}. FVD adapts Frechet Inception Distance (FID) to capture the temporal coherence of a video, while FSIM emphasizes low-level features in IQA metric inspired by the human visual system (HVS). As shown in Table~\ref{table:video_comp}, our model outperforms the other existing approaches in terms of both metrics, demonstrating that our outputs are more similar to the target distribution of the entire video sequences, maintaining per-pixel and structural visual information.

\subsection{Qualitative Analysis}
\label{exp:qualitative}
Further, we qualitatively compare the quality of the restored images with the state-of-the-art approaches: UNet~\citep{ronneberger2015u}, DMCNN~\citep{sun2018moire}, ESDNet~\citep{yu2022towards}, WDNet~\citep{liu2020wavelet}, MBCNN~\citep{zheng2020image}, and VDmoire~\citep{dai2022video}.

\subsubsection{Effect on Restoring Fine-grained Details}
As shown in Fig.~\ref{fig:figure-tcl-result} and~\ref{fig:figure-iphone-result}, we provide (randomly sampled) restored output images as well as a source image (see top left, which clearly has moiré patterns) and a target image (see bottom right). At the bottom of the figure, we provide two magnified image regions (see red and green boxes in the source image) for an effective comparison. Our proposed FPANet shows better image restoration quality, preserving fine-grained details (e.g., sharper edges). Compared with conventional approaches, such as UNet~\citep{ronneberger2015u}, DMCNN~\citep{sun2018moire}, WDNet~\citep{liu2020wavelet}, and MBCNN~\citep{zheng2020image}, which often fail to filter out large-scale moiré patterns, our method effectively removes various sizes of moiré patterns without showing visually obvious artifacts across the image. Importantly, WDNet~\citep{liu2020wavelet} and MBCNN~\citep{zheng2020image} also rely on the frequency domain components using wavelet transform and implicit Discrete Cosine Transform, respectively. This may confirm that our method clearly outperforms the other frequency-based approaches, effectively dealing with various artifacts. We provide more diverse examples in \ref{secA1}.

Further, following \cite{kim2016accurate, dong2016accelerating, mao2016image}, we analyze the image's luminance channel (Y channel in YCbCr color space) to compare the restoration quality, comparing the ability to restore fine-grained features. We also measure PSNR to evaluate the performance on the Y channel. As shown in Fig.~\ref{fig:figure-ychannel}, we visualize a source/target image and restored image patches from existing methods and ours. For better visualization, we only provide an enlarged image patch (see full-size images in the supplemental material). 
Our proposed method recovers the subtle texture details without showing obvious visual artifacts across the frame, which is confirmed by box plots of PSNR values (see Fig.~\ref{fig:figure-ychannel} (right)).

\begin{table*}[t]
    \caption{We provide our ablation study to demonstrate (left) the effect of Frequency Selection Module~(FSM) module while calculating confidence map in terms of each variant and (right) the effect of our Frequency Selection Fusion (FSF) module by replacing it with existing frequency-based building blocks: (i) FFC~\citep{chi2020fast} and (ii) DeepRFT~\citep{mao2021deep}.}
    \label{table:ablationstudy}
    \begin{subtable}[t]{0.44\textwidth}
    \centering
    \label{table:ablation-fsm}
     \resizebox{\textwidth}{!}{
          \begin{tabular}{@{}lcccc@{}} \toprule
            {Methods}  & PSNR$\uparrow$ & SSIM$\uparrow$ & LPIPS$\downarrow$ & Params~(M) \\\midrule
            FSM~(w/ Single Cm) & 21.658  & 0.780 & 0.180 & 68.64 \\
            FSM~(w/o Fusion Cm) & 21.676  & 0.781 & 0.173 & 83.47 \\
            FSM~(w/ Fusion Cm) &  \textbf{21.953} &  \textbf{0.784} & \textbf{0.173}  & 68.89 \\
            \bottomrule
    \end{tabular}
    }
 \end{subtable}
 \hfill
 \begin{subtable}[t]{0.54\textwidth}
 \centering
    %\caption{Ablation studies of FSF Block. }
    \label{table:ablation-fsfblock}
    \resizebox{\textwidth}{!}{
       \begin{tabular}{@{}lcccc@{}} \toprule
            {Methods}  & PSNR$\uparrow$ & SSIM$\uparrow$ & LPIPS$\downarrow$ & Params~(M) \\\midrule
            Ours w/ FFC~\citep{chi2020fast} & 20.893 & 0.744 & 0.199 & 64.57 \\
            Ours w/ DeepRFT~\citep{mao2021deep} & 21.305 & 0.757 & 0.191 & 57,46 \\\midrule
            Ours w/ FSF &  \textbf{21.953} &  \textbf{0.784} & \textbf{0.173} & 68.89 \\
            \bottomrule
    \end{tabular}
    }
 \end{subtable}
\end{table*}

\subsubsection{Robustness against Color Degradation}
ESDNet~\citep{yu2022towards} and VDmoire~\citep{dai2022video} also provide a compelling quality demoiréing, but we observe VDmoire~\citep{dai2022video} often suffer from color shifts, probably due to a lack of the model's color restoration power. VDmoire~\citep{dai2022video} depends on the pixel's statistical information (i.e. mean and variance) across the multiple consecutive frames for temporal consistency, which makes it difficult to restore accurate pixel value per frame, resulting in making toned-down images. However, ours, which uses the Frequency Spatial Fusion (FSF) module, shows fewer artifacts in color degradation. We also observe that ESDNet~\citep{yu2022towards} often generates images with low visual acuity, failing to restore the original visual contents. This is more apparent in examples shown in Fig.~\ref{fig:figure-iphone-result} (see some blurry characters, which is not the case for ours). These confirm that our proposed method effectively deals with moiré restoration with fewer artifacts, such as color degradation, lack of sharpness, and remaining large-scale moiré patterns. We provide more diverse examples in \ref{secA1}.

Further, we also compare RGB color histograms with VDmoire to compare the amount of color degradation as shown in Fig.~\ref{fig:figure-color-histogram} (a). To quantify the color degradation, we measure the average correlation (for each channel) with those of the target (clear) image. A source image with moiré patterns shows the smallest correlation value, 0.602, while our proposed method shows the highest correlation value, 0.746, which is better than VDmoire. In Fig.~\ref{fig:figure-color-histogram} (b), we provide box plots of such correlations for all test images, further confirming that our proposed method is more robust to color degradation.

\subsection{Ablation Study}
\subsubsection{Effect of Individual Modules}
Our proposed model consists of four main components: (i) Amplitude and Phase, (ii) FSM (Frequency Selection Module), (iii) CSFM (Cross Scale Fusion Module), and (iv) PAM (Post Align Module).
To analyze their contribution, we conduct an ablation experiment with various combinations of several building blocks based on Model 1, which utilizes the UNet architecture composed of residual block~\citep{he2016deep}.
Note that we use the TCL-V1 dataset for this experiment. 
We observe in Table~\ref{table:ablation-module} that each building component equally improves the overall image demoiréing performance in terms of PSNR and SSIM. For example, removing each building block decreases the overall performance (compare Model 6 vs. Model 2, 3, 4, and 5). 
Specifically, we notice that the comparison results of Model 2, Model 3, and Model 4 demonstrate that the performance gain mainly stems from each frequency component and Frequency Selection Module~(FSM) design rather than the increase of parameters.

\subsubsection{Effect of Loss Functions}
Recall from Sect.~\ref{sec:loss}; we train our model with the following three loss terms: (i) Multi-scale spatial domain loss $\mathcal{L}_s$, (ii) Multi-scale perceptual loss $\mathcal{L}_\text{vgg}$, and (iii) Frequency domain loss $\mathcal{L}_\text{freq}$. We also conduct an ablation study with different combinations of the loss terms to see their individual impacts. 
As shown in Table~\ref{table:ablation_loss}, we observe all loss terms have noticeable effects on the overall performance, and using all loss terms provides the best performance in all three metrics.
The multi-scale perceptual loss function especially helps generate more realistic images. 

\begin{figure}[hbt!]
    \centering
    % \vspace{-1cm}
    \includegraphics[width=\linewidth]{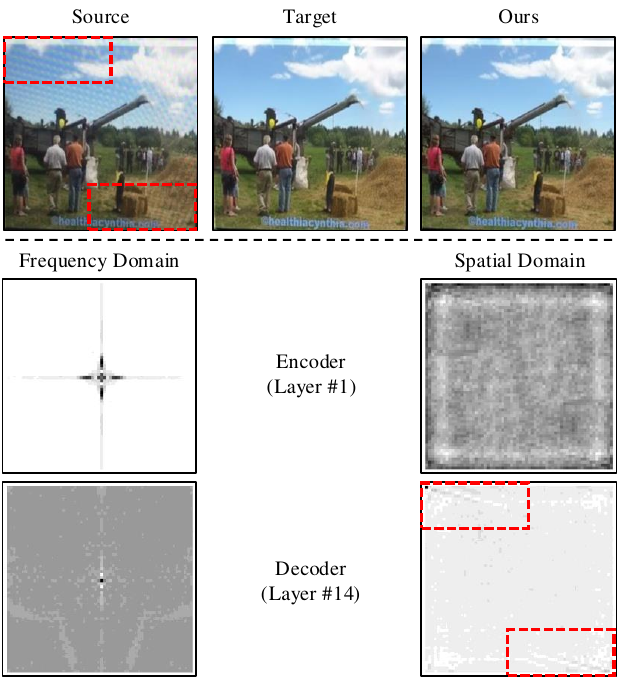}
    \caption{Visualize the features in both the frequency and spatial domains obtained from the Frequency Selection Module (FSM). The red box highlights the specified region of interest.}
    \label{fig:review_2}
\end{figure}

\subsubsection{Effect of Frequency Spatial Fusion (FSF) block}
We propose a novel encoding module called the Frequency Spatial Fusion module, which utilizes frequency domain components and geometric structural information. In this section, we present a deep analysis of FSF module.

\noindent\textbf{Analysis on Frequency Selection Module~(FSM)} In Sect.~\ref{sec:fsm}, we apply a confidence map that comes from a combined feature map on amplitude and phase respectively to enhance the effectiveness of utilizing two related properties.
In Table~\ref{table:ablationstudy}~(left), each variant of FSM is followed by: (i) $1^\text{st}$ row~(w/ Single Cm) indicates that adopts only one confidence map for both amplitude and phase and (ii) $2^\text{nd}$ row~(w/o Fusion Cm) denotes that performs a confidence map in terms of each property without the fusion stage.
We observe that the proposed combined confidence map leveraging each strongly correlated property, amplitude, and phase, brings a performance gain in all metrics.

\noindent\textbf{Visualization of selected frequency bands} We also provide the visualization of features obtained from the Frequency Selection Module~(FSM). To visualize the frequency domain feature, we select the top 30 channels extracted from FSM that exhibit the highest correlation with the content. Then, we apply a channel-wise average pooling to the obtained feature map.
Additionally, to obtain the spatial domain feature map, we apply the inverse Fourier transform using the acquired frequency and phase features. As shown in the Fig.~\ref{fig:review_2}, we observe that features extracted from the early layers capture information with regard to the entire content. 
On the other hand, features from the final layer emphasize highly moiré-correlated frequency bands, indicating that our proposed FSM plays a crucial role in disentangling moiré patterns from the original contents in the frequency domain.

\noindent\textbf{Effect of Frequency Spatial Fusion~(FSF)} To further demonstrate the effectiveness of our FSF module, we replace it with existing frequency-based encoding modules: FFC~\citep{chi2020fast} and DeepRFT~\citep{mao2021deep}, which directly use the frequency components with spatial domain features instead of dealing with amplitude and phase components separately.
As shown in Table~\ref{table:ablationstudy} (right), we report the results over multi-frame experiments and number of the parameters. Although our full architecture has more parameters than others, it is clear that amplitude and phase properties lead to performance gain on demoiré through comparison with Model 3 and Model 4 in Table~\ref{table:ablation-module} rather than directly using frequency feature.

\begin{figure*}[t!]
    \centering
    \includegraphics[width=\textwidth]{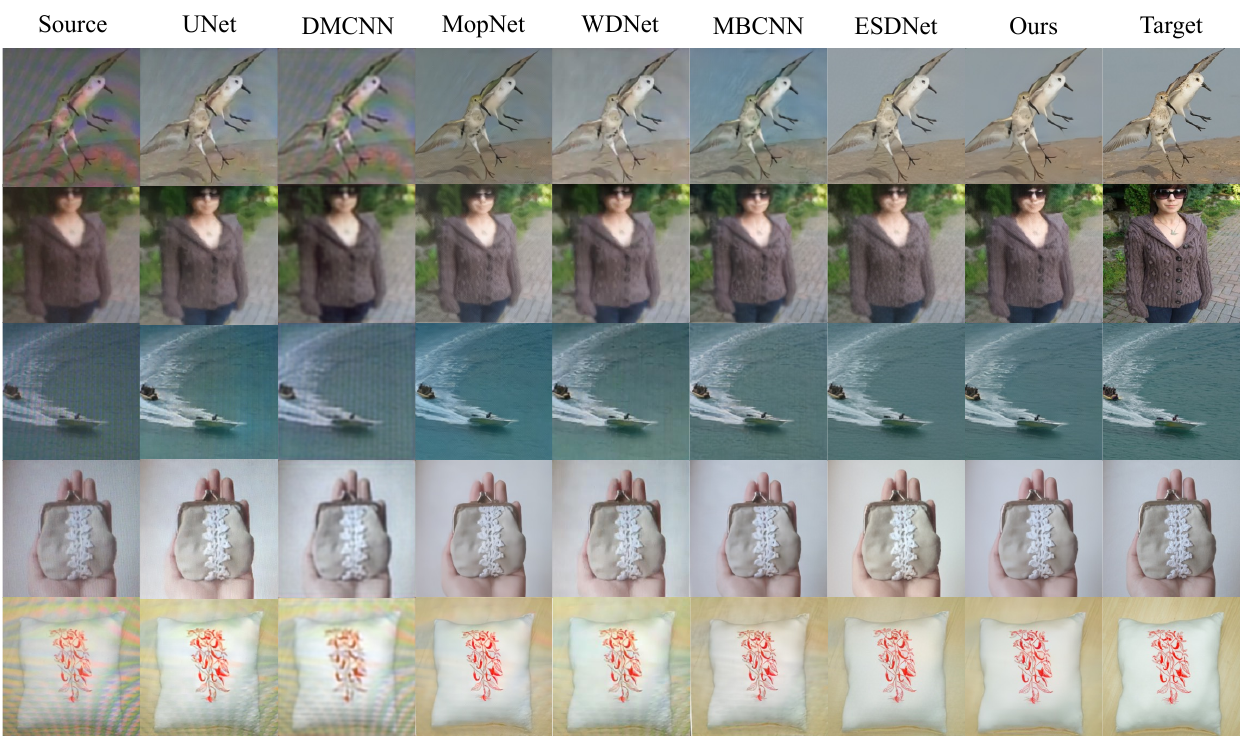}
    \caption{Qualitative comparison on TIP2018 dataset~\citep{sun2018moire}. Please zoom in for details.}
    \label{fig:tip2018}
\end{figure*}

\begin{table}
\begin{center}
\caption{We provide our ablation study to demonstrate the effect of our Post Align Module (PAM) module, where E and D indicate the location of PAM (i.e., encoder and decoder stage).}
\label{table:pam_ablation}
 \resizebox{\linewidth}{!}{%
        \begin{tabular}{@{}lcccc@{}} \toprule
                {Methods}  & PSNR$\uparrow$ & SSIM$\uparrow$ & LPIPS$\downarrow$ & Params~(M) \\\midrule
                Ours w/ PCD~\citep{wang2019edvr} & 21.507  & 0.780 & 0.184 & 64.53 \\
                Ours w/ PAM~(E) & 21.806  & 0.784 & 0.175 & 69.54 \\
                Ours w/ PAM~(D) &  \textbf{21.953} &  \textbf{0.784} & \textbf{0.173}  & 68.89 \\
                \bottomrule
        \end{tabular}}
\end{center}
\end{table}

\subsubsection{Effect of Temporal Feature Alignment by PAM}
We further compare our temporal feature alignment method Post Align Module (PAM) with existing Pyramid Cascade Deformable technique~\citep{wang2019edvr}, which is utilized in VDmoire~\citep{dai2022video} to extract implicitly aligned features between consecutive frames. As we summarized in Table~\ref{table:pam_ablation}, our proposed method outperforms the alternative in all metrics, including PSNR, SSIM, and LPIPS: 0.45 dB, 0.004, and 0.009 gains in PSNR, SSIM, and LPIPS, respectively. 
Additionally, we report the performance when the position of PAM is changed from the decoder to the encoder. 
The difference between PCD and pre-align is that pre-align combines the auxiliary features from nearby frames inside the main architecture, whereas PCD does so before the main architecture.
The results demonstrate that post-align technique is effective in demoiré task especially in pixel-level distortion (PSNR) while maintaining perceptual quality (SSIM and LPIPS) compared with pre-align.

\begin{table}
\begin{center}
    
  \caption{Quantitative comparison with the state-of-the-art image demoiréing approaches on TIP2018 dataset~\citep{sun2018moire}.}
  \label{tab:tip2018_table}
  \resizebox{\linewidth}{!}
  {
  \begin{tabular}{cc|c|c|c|c }
  \toprule
      &Method & PSNR$\uparrow$ & SSIM$\uparrow$ &LPIPS$\downarrow$ &Params~(M)\\
    \hline
    &UNet~\citep{ronneberger2015u}&$26.49$&$0.864$&$0.432$& 2.35\\
    &DMCNN~\citep{sun2018moire}&$26.77$&$0.871$ &$0.341$& 1.43\\
    &MopNet~\citep{He_2019_ICCV}&$27.75$&$0.895$ &$0.378$& 58.57\\
    &WDNet~\citep{liu2020wavelet}&$28.08$&$0.904$ &$0.213$& 3.36\\
    &MBCNN~\citep{zheng2020image}&$30.03$&$0.893$ &$0.095$& 14.19\\
    &ESDNet~\citep{yu2022towards}&$30.11$&$0.920$ &$0.072$& 10.62\\
    \hline
    &Ours &\textbf{30.46}&\textbf{0.923}&\textbf{0.070} & 21.72\\
    \bottomrule[0.1em]
  \end{tabular}}
\end{center}
\end{table}

\subsubsection{Demoiré in Image Domain}
\noindent\textbf{Analysis on TIP2018 dataset} To validate the ability of our proposed method in the demoiré task, especially the FSF module, which considers amplitude and phase properties simultaneously, we extend the experiments into the image domain. 
We train our model with an image demoiré benchmark dataset, called TIP2018~\citep{sun2018moire}, which consists of 135,000 image pairs. 
Further, we apply a similar setting with a single-frame experiment except for the number of blocks. For the image domain experiment, we change the number of blocks from [2, 2, 4] to [2, 2, 2] in each stage and 12 to 2 in the bottleneck.
Table~\ref{tab:tip2018_table} shows the demoiré results compared with the existing image demoiré approaches. Our proposed method achieves the best scores in PSNR, SSIM, and LPIPS, 30.46 dB, 0.923, and 0.070, which are higher than those of the previous state-of-the-art methods.
Moreover, we present a qualitative comparison in Fig.~\ref{fig:tip2018} with other competing methods. The recovered images produced by our method demonstrate that the moiré pattern is well removed, and they are visually closer to the ground truth.

\begin{figure*}[hbt!]
    \centering
    \includegraphics[width=\textwidth]{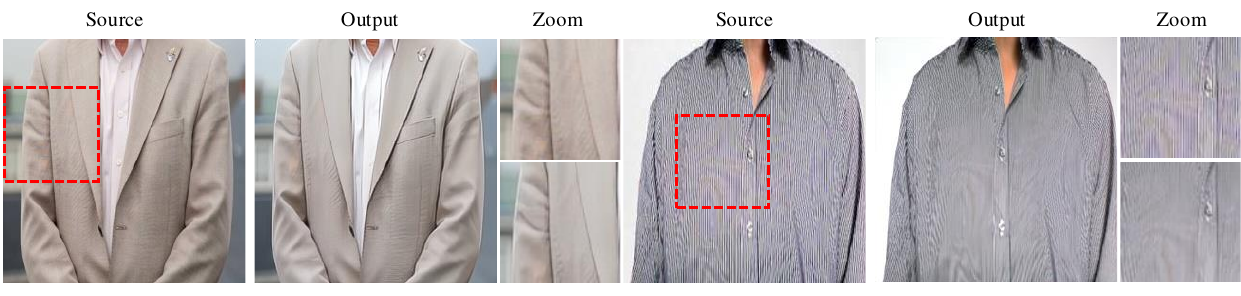}
    \caption{Visualization of the output using web-crawled images showing moiré patterns on fabrics. Red box indicates the zoomed area.}
    \label{fig:fabric}
\end{figure*}

\noindent\textbf{Analysis on non-screen-captured moiré image} We also conduct additional experiments to validate the effectiveness of our methods on non-screen-captured moiré degraded images. Since there are no publicly available datasets for the experiment, we manually collect the texture-degraded images from the web. Although the apples to apples comparison is not possible due to the absence of ground truth, our results demonstrate that moiré patterns present on fabrics are effectively removed, as clearly indicated in Fig.~\ref{fig:fabric}.

\subsection{Future Works}
Although our proposed approach demonstrates effectiveness in removing moiré patterns with comprehensive understanding in frequency domain, we utilize simple multi-scale features in the spatial domain. We recommend further research to extend the task by incorporating the advanced feature fusion pipeline to enhancing spatial features. Furthermore, we propose enhancing the capabilities of amplitude and phase characteristics through a sophisticated integration with spatial domain features.

\section{Conclusion}\label{sec5}
We introduced a novel video demoiréing method called FPANet by proposing the following two components: (i) Frequency Selection Fusion (FSF) and (ii)  Post Align Module (PAM).
In proposed FSF block, FSM utilizes amplitude and phase components in the frequency domain to address undesired color changes and large-scale moiré patterns, while CSFM is used to capture multi-scale features to recover both global and fine-grained features. 
Lastly, PAM is utilized to align features from multiple consecutive frames with reduced visual artifacts.
We demonstrated the effectiveness of using our proposed method with a public video demoiréing dataset called VDmoire, and ours generally outperforms existing state-of-the-art approaches in terms of various image and video quality metrics, such as PSNR, SSIM, LPIPS, FVD, and FSIM.

\section*{Acknowledgements}
This work was supported by LG Display Research Center 50\%. This work was also supported by Institute of Information \& communications Technology Planning \& Evaluation (IITP) grant funded by the Korea government(MSIT) (2022-0-00043, 15\%, IITP-2024-2020-0-01819, 10\%),
the National Research Council of Science \& Technology (NST) grant by the Korea government (MSIT) (No. CRC21011, 16\%),
Culture, Sports, and Tourism R\&D Program through the Korea Creative Content Agency grant funded by the Ministry of Culture, Sports and Tourism in 2024 (International Collaborative Research and Global Talent Development for the Development of Copyright Management and Protection Technologies for Generative AI, RS-2024-00345025, 8\%),
and Institute of Information \& communications Technology Planning \& Evaluation (IITP) grant funded by the Korea government(MSIT) (No. 2019-0-00079, Artificial Intelligence Graduate School Program(Korea University), 1\%).

\section*{Author contributions}
Conceptualization: Gyeongrok Oh, Heon Gu, Jinkyu Kim, Sangpil Kim; Methodology: Gyeongrok Oh, Jinkyu Kim, Sangpil Kim; Formal analysis and investigation: Gyeongrok Oh, Sungjune Kim, Sang Ho Yoon, Heon Gu, Jinkyu Kim, Sangpil Kim; Writing - original draft preparation: Gyeongrok Oh, Sungjune Kim, Jinkyu Kim, Sangpil Kim; Supervision: Jinkyu Kim, Sangpil Kim;

\appendix

\renewcommand\thefigure{\arabic{figure}}    
\setcounter{figure}{15}    
%%=============================================%%
%% For submissions to Nature Portfolio Journals %%
%% please use the heading ``Extended Data''.   %%
%%=============================================%%

%%=============================================================%%
%% Sample for another appendix section			       %%
%%=============================================================%%

\section{More Qualitative Results}\label{secA1}
%% Appendices may be used for helpful, supporting or essential material that would otherwise 
%% clutter, break up or be distracting to the text. Appendices can consist of sections, figures, 
%% tables and equations etc.

\begin{figure*}[!]
    \centering
    % \vspace{-1cm}
    \includegraphics[width=\textwidth]{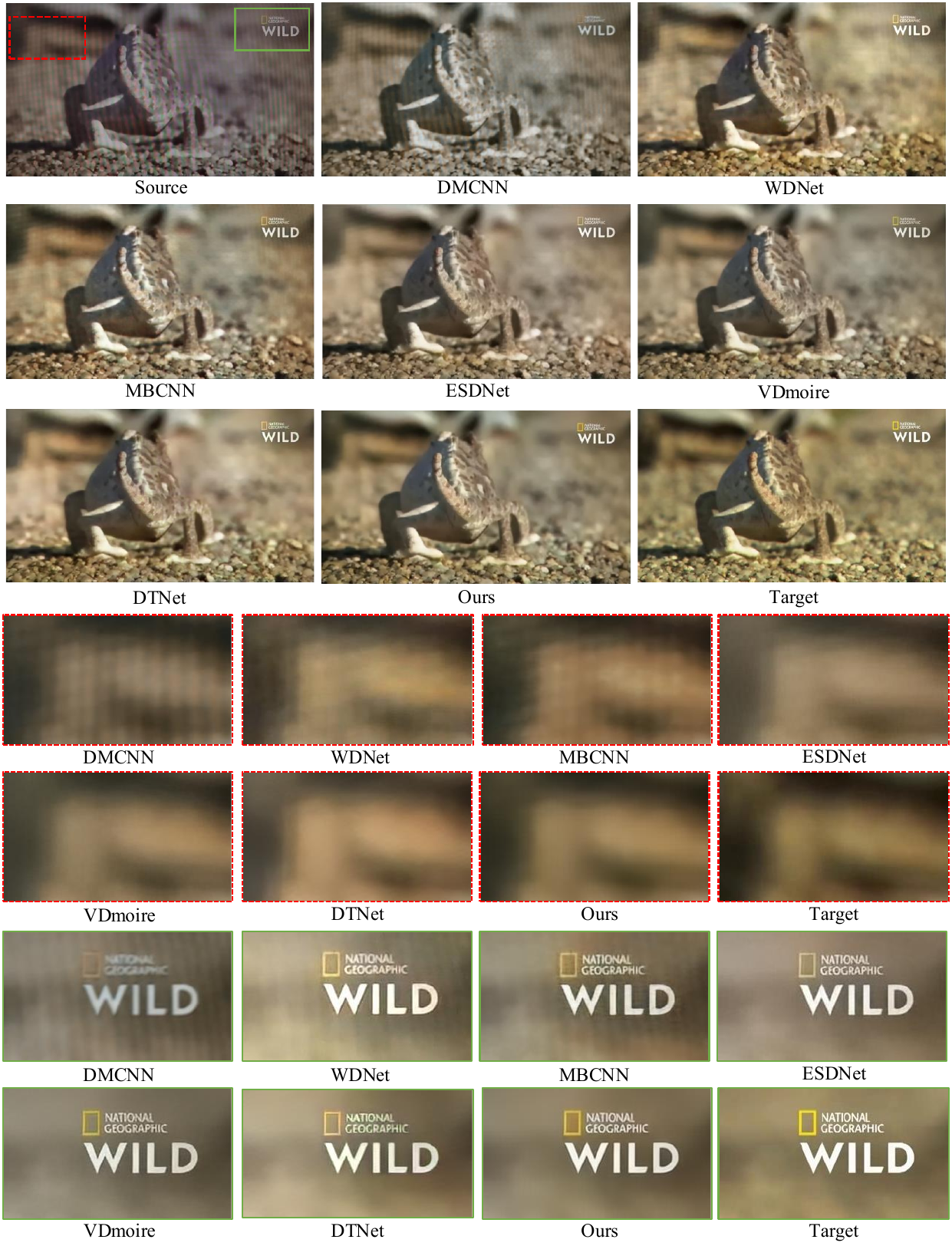}
    \caption{Qualitative comparison on the TCL-V1. The red and green boxes zoom in on frames to obviously compare the results.}
    \label{fig:figure_app_result_1}
\end{figure*}
% \clearpage

\begin{figure*}[!]
    \centering
    % \vspace{-1cm}
    \includegraphics[width=\textwidth]{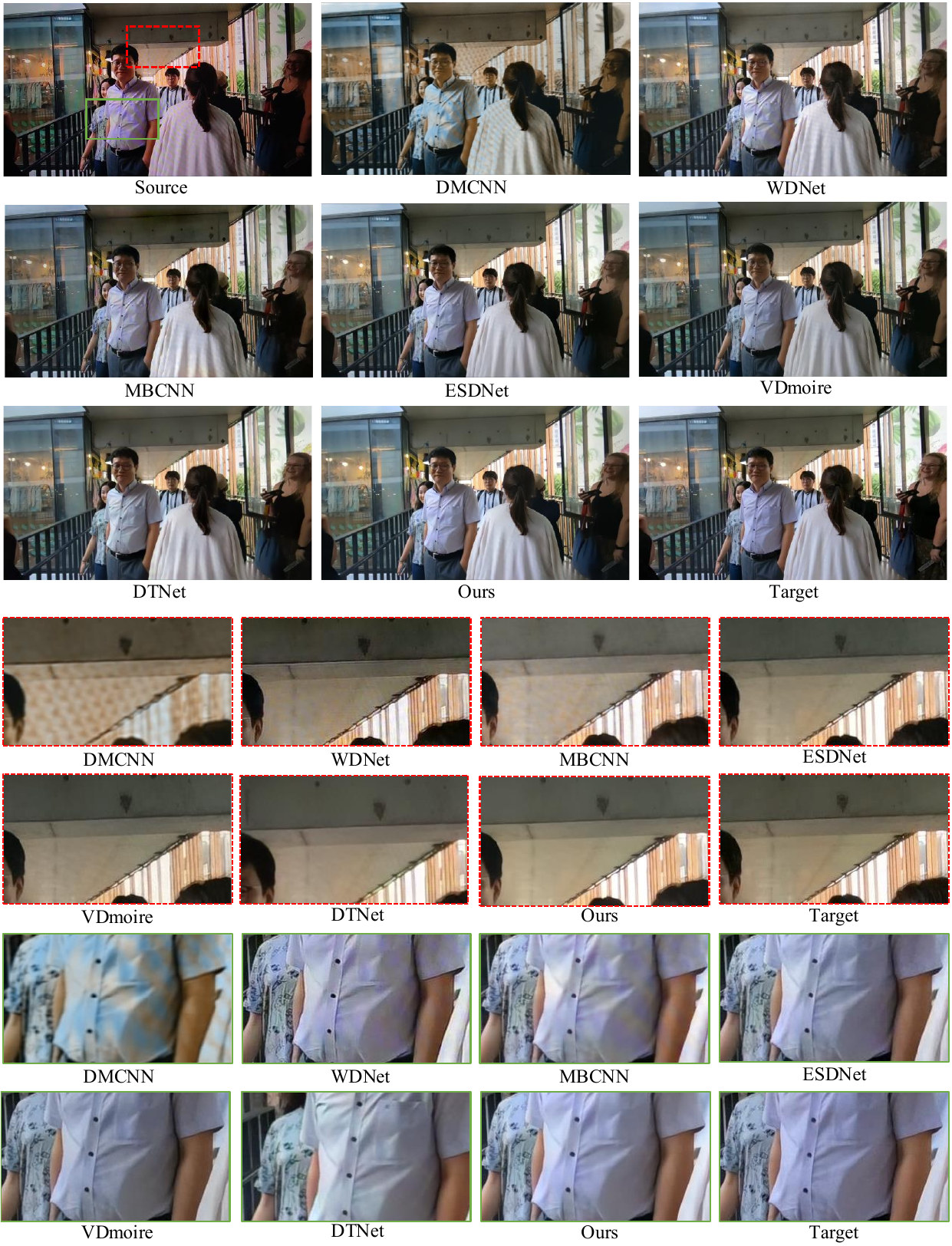}
    \caption{Qualitative comparison on the TCL-V1. The red and green boxes zoom in on frames to obviously compare the results.}
    \label{fig:figure_app_result_2}
\end{figure*}

\begin{figure*}[!]
    \centering
    % \vspace{-1cm}
    \includegraphics[width=\textwidth]{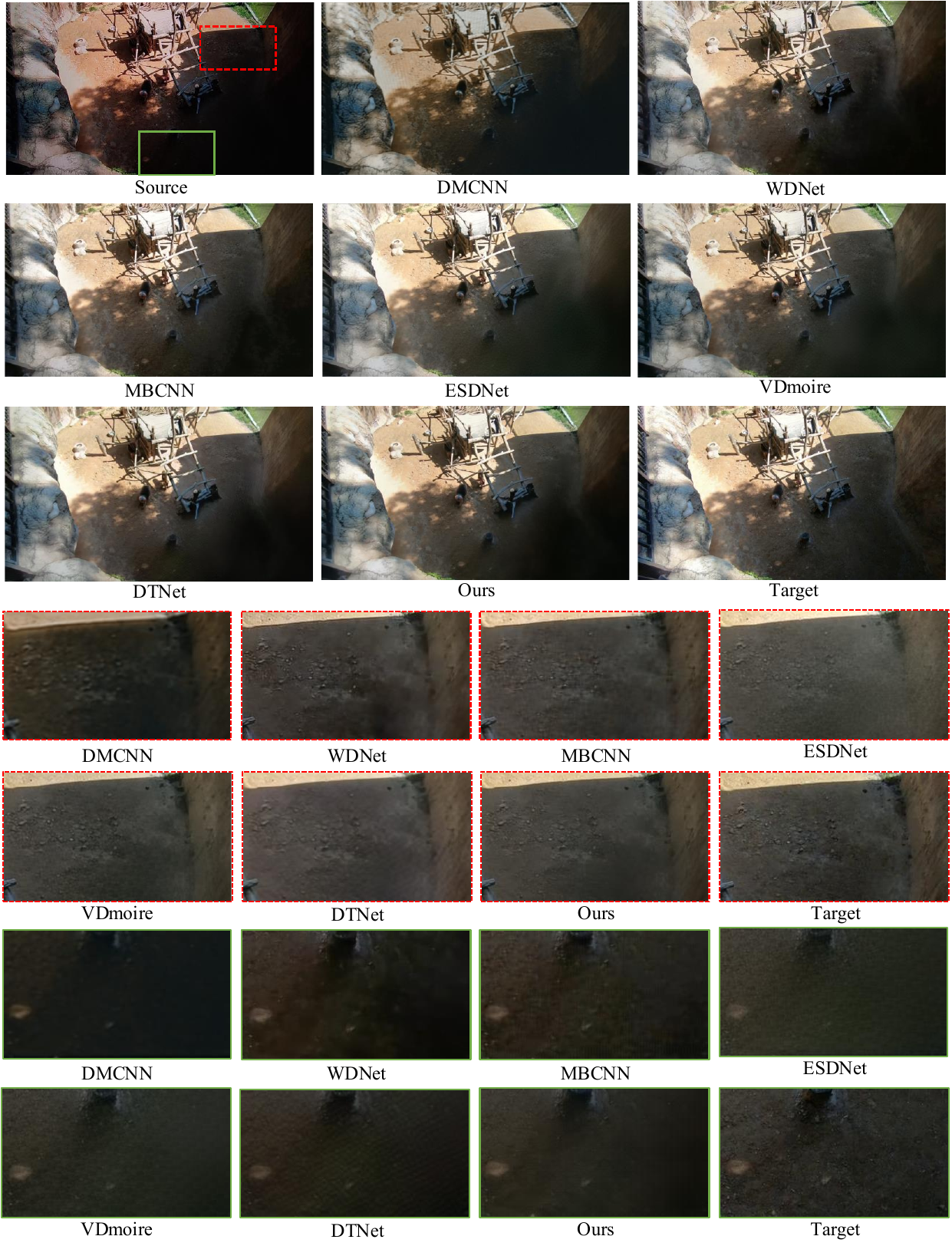}
    \caption{Qualitative comparison on the TCL-V1. The red and green boxes zoom in on frames to obviously compare the results.}
    \label{fig:figure_app_result_3}
\end{figure*}

\begin{figure*}[!]
    \centering
    % \vspace{-1cm}
    \includegraphics[width=\textwidth]{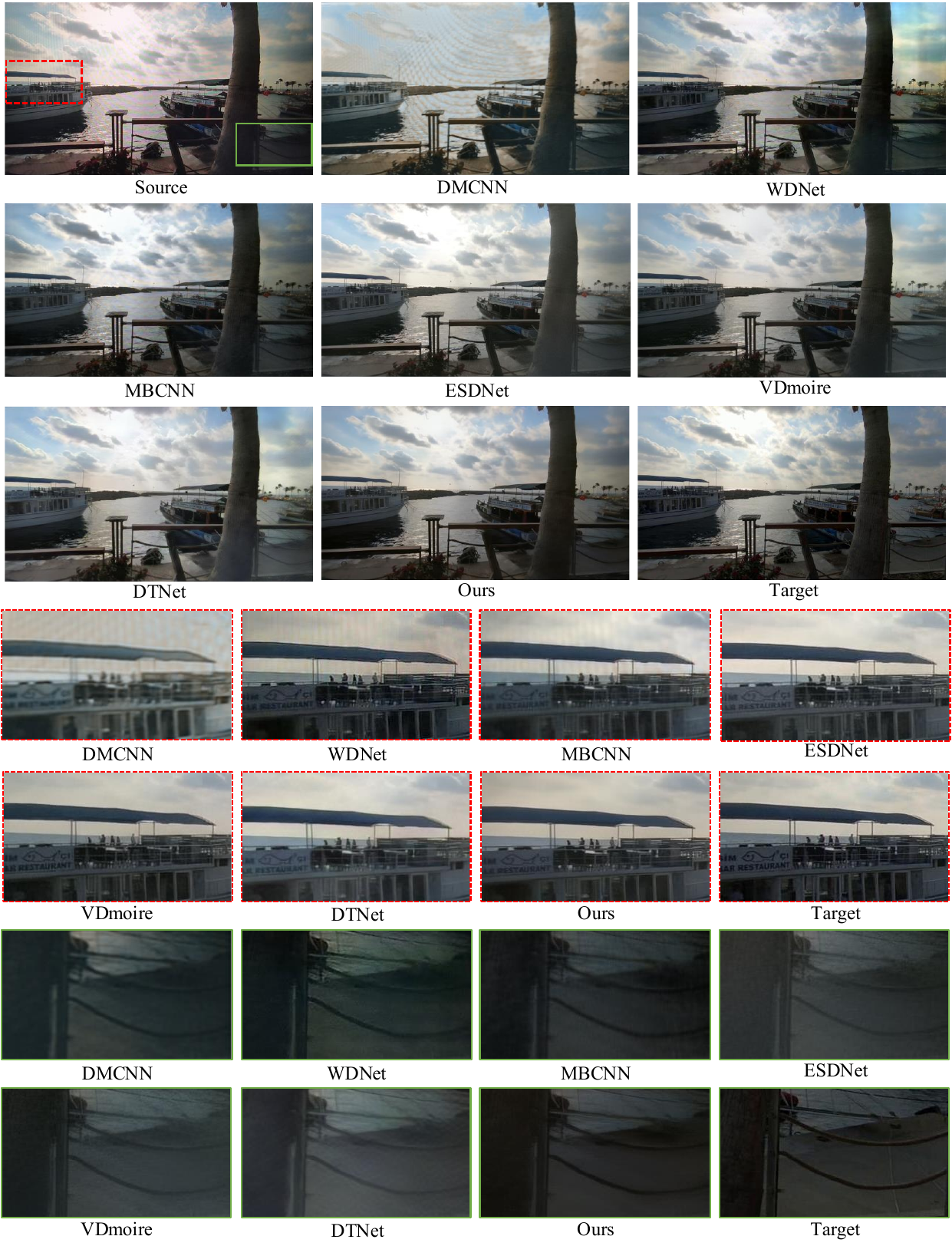}
    \caption{Qualitative comparison on the TCL-V1. The red and green boxes zoom in on frames to obviously compare the results.}
    \label{fig:figure_app_result_4}
\end{figure*}

\begin{figure*}[!]
    \centering
    % \vspace{-1cm}
    \includegraphics[width=\textwidth]{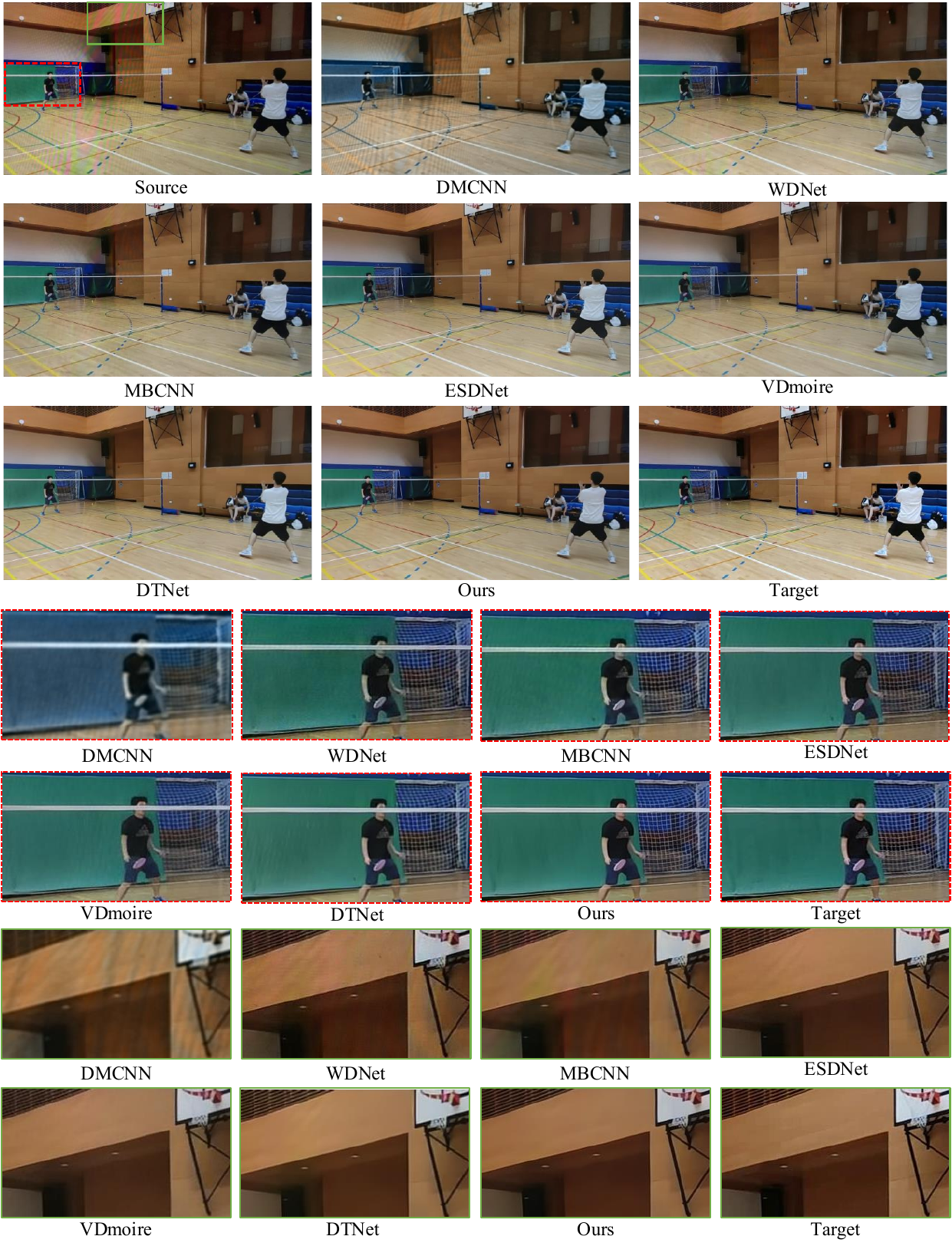}
    \caption{Qualitative comparison on the iPhone-V2. The red and green boxes zoom in on frames to obviously compare the results.}
    \label{fig:figure_app_result_5}
\end{figure*}
\clearpage

%% If you have bibdatabase file and want bibtex to generate the
%% bibitems, please use
%%
% \bibliographystyle{elsarticle-harv} 
\bibliographystyle{abbrvnat} 
\bibliography{main}

%% else use the following coding to input the bibitems directly in the
%% TeX file.

% \begin{thebibliography}{00}

% %% \bibitem[Author(year)]{label}
% %% Text of bibliographic item

% \bibitem[ ()]{}

% \end{thebibliography}
\end{document}